\documentclass[11pt]{article}

\usepackage[margin=2.0cm]{geometry}
\usepackage{amsmath,amssymb,amsfonts}
\usepackage{graphicx}
\usepackage{bm}
\usepackage{xcolor}
\usepackage{caption}
\usepackage[numbers,super,sort&compress]{natbib}
\usepackage{hyperref}
\hypersetup{colorlinks=true,linkcolor=blue,citecolor=blue,urlcolor=blue}

\usepackage{float}
\usepackage{booktabs}
\usepackage{tabularx}
\usepackage{array}
\usepackage{pifont}

\newcommand{\suppfigtoc}[2]{%
    \phantomsection
    \addcontentsline{toc}{subsection}{%
        Supplementary Figure~\ref{#1}. #2}%
}
\newcommand{\supptabtoc}[1]{%
    \phantomsection
    \addcontentsline{toc}{subsection}{%
        Supplementary Table~\thetable: #1}%
}

\definecolor{walkcolor}{rgb}{0.2, 0.2, 0.2}
\definecolor{walk2color}{rgb}{0.7, 0.7, 0.7}
\definecolor{amblecolor}{rgb}{0.0, 0.4470, 0.7410}
\definecolor{trotcolor}{rgb}{0.8500, 0.3250, 0.0980}
\definecolor{pacecolor}{rgb}{0.9290, 0.6940, 0.1250}
\definecolor{pronkcolor}{rgb}{0.4660, 0.6740, 0.1880}
\definecolor{halfboundcolor}{rgb}{0.9490, 0.4030, 0.7720}
\definecolor{boundcolor}{rgb}{0.4940, 0.1840, 0.5560}
\definecolor{cantercolor}{rgb}{0.3010, 0.7450, 0.9330}
\definecolor{tgallopcolor}{rgb}{0.6350, 0.0780, 0.1840}
\definecolor{rgallopcolor}{rgb}{0.1, 0.8, 0.5}

\newcommand{\ie}{\textit{i.e.}}

\title{Learning optimal gaits and styles for terrain-adaptive quadruped locomotion}

\author{
  Guillaume Bellegarda$^{1,2*}$ and Auke Ijspeert$^{2}$ \\[6pt]
  \small $^{1}$Inria, Université de Lorraine, CNRS, 54000 Nancy, France \\
  \small $^{2}$École Polytechnique Fédérale de Lausanne, 1015 Lausanne, Switzerland\\
  \footnotesize $^{*}$Corresponding author: guillaume.bellegarda@inria.fr
}

\date{}

\begin{document}
\maketitle

\begin{abstract}

Animals use many different gaits to locomote at different speeds across varying terrains. A longstanding question in biology and robotics is: which gait is best? This question is most often addressed from an energy-efficiency perspective on flat terrain, leaving open how the answer depends on terrain and on the criterion optimized. Here we show that gait optimality is determined jointly by terrain, performance metric, and velocity. We use deep reinforcement learning and networks of abstract oscillators, modeling spinal central pattern generators, to learn 11 quadrupedal gaits across diverse challenging terrains, together with the locomotion style (base height, foot ground clearance, foot offsets) that each requires. A second policy then selects among them to optimize a user-specified criterion such as cost of transport, base stability, or joint smoothness. Across these performance metrics and within this repertoire, every gait is optimal under at least one terrain--metric--velocity combination. The framework adapts zero-shot to gait couplings and styles never seen during training, to terrains far beyond the training distribution, and to larger quadrupeds five times the training mass. With a single additional pitch-tracking reward term, the same gait couplings also support bipedal-mode locomotion. Our method provides a general and systematic way to produce expressive locomotion and to select the gait best suited to a given terrain and criterion. Project page: \url{https://guillaume10.github.io/terrain-gaits}.

\end{abstract}

\section*{Introduction}
\label{sec:intro}

The remarkable locomotor versatility of quadrupedal animals, smoothly transitioning from gaits such as walk to trot to gallop depending on speed, terrain, and metabolic cost, has captivated biologists and engineers alike for decades\cite{hildebrand1965symmetrical,hildebrand1977analysis,alexander1984gaits,grillner2020current}. A central question in both communities is deceptively simple: \emph{which gait is best?} In biology, this question is most often studied from an energy-efficiency perspective on flat terrain, with a central hypothesis that gaits are selected to minimize metabolic cost\cite{hoyt1981gait}. Yet animals routinely negotiate diverse terrains including slopes, gaps, rubble, and stairs, where the answer almost certainly depends on more than energy alone. Recent work has shown that viability (i.e.~fall avoidance) may be a more fundamental driver of gait transitions than energy efficiency, particularly on challenging terrain\cite{shafiee2024viability}. Stability, energy efficiency, and robustness all matter, and may favor different gaits under different conditions.

In robotics, recent advances in deep reinforcement learning (DRL) have produced quadrupedal controllers capable of agile parkour\cite{cheng2024extreme,zhuang2023robot,hoeller2024anymal}, robust locomotion over natural terrain\cite{lee2020learning,miki2022learning}, and high-speed running\cite{bellegarda2022robust,margolis2022rapid,ji2022concurrent}. In these systems, the gait \emph{emerges} implicitly through training, shaped by the reward function and curriculum\cite{fu2022energy}: most commonly a trot for flat-ground locomotion~\cite{ji2022concurrent,margolis2022rapid,miki2022learning,rudin2022anymalisaac}. In parkour settings, these controllers prioritize obstacle traversal over tracking a range of user-specified velocities: the gait is also not an independent user command, and the resulting contact pattern and speed are largely outcomes of training\cite{cheng2024extreme,zhuang2023robot}. Gait-specific approaches have trained separate policies for trotting, bounding, pacing, and galloping\cite{yu2023identifying}, prescribed bounding for continuous stair jumping\cite{yang2024agile}, or imitated animal gaits from motion-capture data\cite{peng2020learning,escontrela2022adversarial,han2024lifelike,fu2026learning}. While imitation learning produces naturalistic behaviors, the available gait repertoire depends on the reference motions, whose joint trajectories need not be optimal for a robot with a different morphology. Across these approaches, gait performance is not systematically compared over a broad repertoire under varying terrain, velocity, and performance criteria, leaving the relative optimality of individual gaits unclear.

From the model-based perspective, model-predictive control (MPC) methods for quadruped locomotion typically include a fixed or variable-frequency gait schedule as part of the optimization\cite{dicarlo2018mpc,kim2019highly,bellicoso2018dynamic}. When considering rough terrain, gait sequences can be co-optimized with foothold patterns and body trajectories\cite{winkler2018gait,aceituno2018simultaneous}, while footholds can also be optimized for a given gait\cite{jenelten2022tamols}, and neural networks can be trained to track such optimized trajectories as online feedback policies\cite{jenelten2024dtc}. Other approaches learn gait transitions primarily as a function of speed\cite{yang2022fast}, use offline trajectory optimization to inform gait selection\cite{wang2020automatic}, or constrain predefined gait choices using terrain classification\cite{luo2025tfgait}. Hybrid learning and model-based controllers have also explored fixed and adaptive contact schedules for jumping tasks, including pronking and bounding\cite{margolis2022pixels,yang2023cajun}. Despite these advances, gait-sequence optimization is often computationally demanding or performed offline, while online gait selection is typically limited to a small set of candidates. The combinatorial problem of evaluating many gaits across diverse terrains and metrics thus remains largely unaddressed.

In animals, central pattern generators (CPGs) are neural circuits in the spinal cord of vertebrates that produce rhythmic motor patterns for locomotion~\cite{ijspeert2008,grillner2020current}. The diversity of quadrupedal gaits, from symmetrical gaits like walk and trot to asymmetrical gaits like canter and gallop, has been systematically classified by Hildebrand\cite{hildebrand1965symmetrical,hildebrand1977analysis}, and this taxonomy forms the basis for the 11 gaits considered in this work. In robotics, CPGs provide a principled way to encode gait patterns through coupled oscillators that enforce prescribed phase relationships. The CPG-RL framework\cite{bellegarda2022cpgrl} integrated networks of  oscillators into deep reinforcement learning, allowing the policy to modulate their intrinsic amplitude and frequency setpoints to generate coordinated task-space foot trajectories, with the internal CPG states serving as an efference copy\cite{von1950reafferenzprinzip,crapse2008corollary} providing noise-free gait phase information to the policy. This was extended to visually-guided obstacle avoidance, leveraging explicit inter-oscillator coupling for robust sim-to-real transfer\cite{bellegarda2024visualcpgrl}, and to challenging terrain, where gait transitions emerged as a means of maintaining viability\cite{shafiee2024viability}. Other oscillator-based approaches have explored spontaneous gait emergence through local feedback\cite{owaki2017quadruped,zhang2024learning}, while online optimization has been used to adapt to changing terrain and payloads\cite{zhang2024online}, and to optimize continuous jumping motions directly on hardware\cite{bellegarda2024quadruped}.  Subsequently, AllGaits\cite{bellegarda2025allgaits} showed that nine quadrupedal gaits and transitions between them could be learned with a single policy, but was limited to flat-terrain gait optimality with fixed locomotion styles. Other multi-gait learning approaches condition policies on explicit gait parameters\cite{margolis2022walktheseways}, learn gait transitions via phase-guided controllers\cite{shao2022gait} or latent representations\cite{liu2025skill,mitchell2025gaitor}, and investigate energy-efficient gait transitions or biomechanics-inspired gait selection\cite{stamatopoulou2025egait,humphreys2025learning}. Concurrent work has also demonstrated perceptive selection between learned trotting and bounding skills using trajectory-optimization-derived motion priors\cite{kang2026agile}. Despite these advances, a key open question is how to learn to execute a broad repertoire of gaits with terrain-adaptive locomotion styles and systematically evaluate their performance across terrains, velocities, and performance metrics.

To the best of our knowledge, no existing framework can simultaneously (i)~learn to execute a broad repertoire of quadrupedal gaits across diverse challenging terrains, (ii)~co-optimize gait parameters and locomotion style for each gait and terrain, and (iii)~systematically compare gait performance under multiple competing criteria. Such a framework would let us move beyond asking \emph{which gait emerged?}\ to asking \emph{which gait \textbf{should} we use, and why?}

\begin{figure*}[!t]
    \centering
    \includegraphics[width=\textwidth]{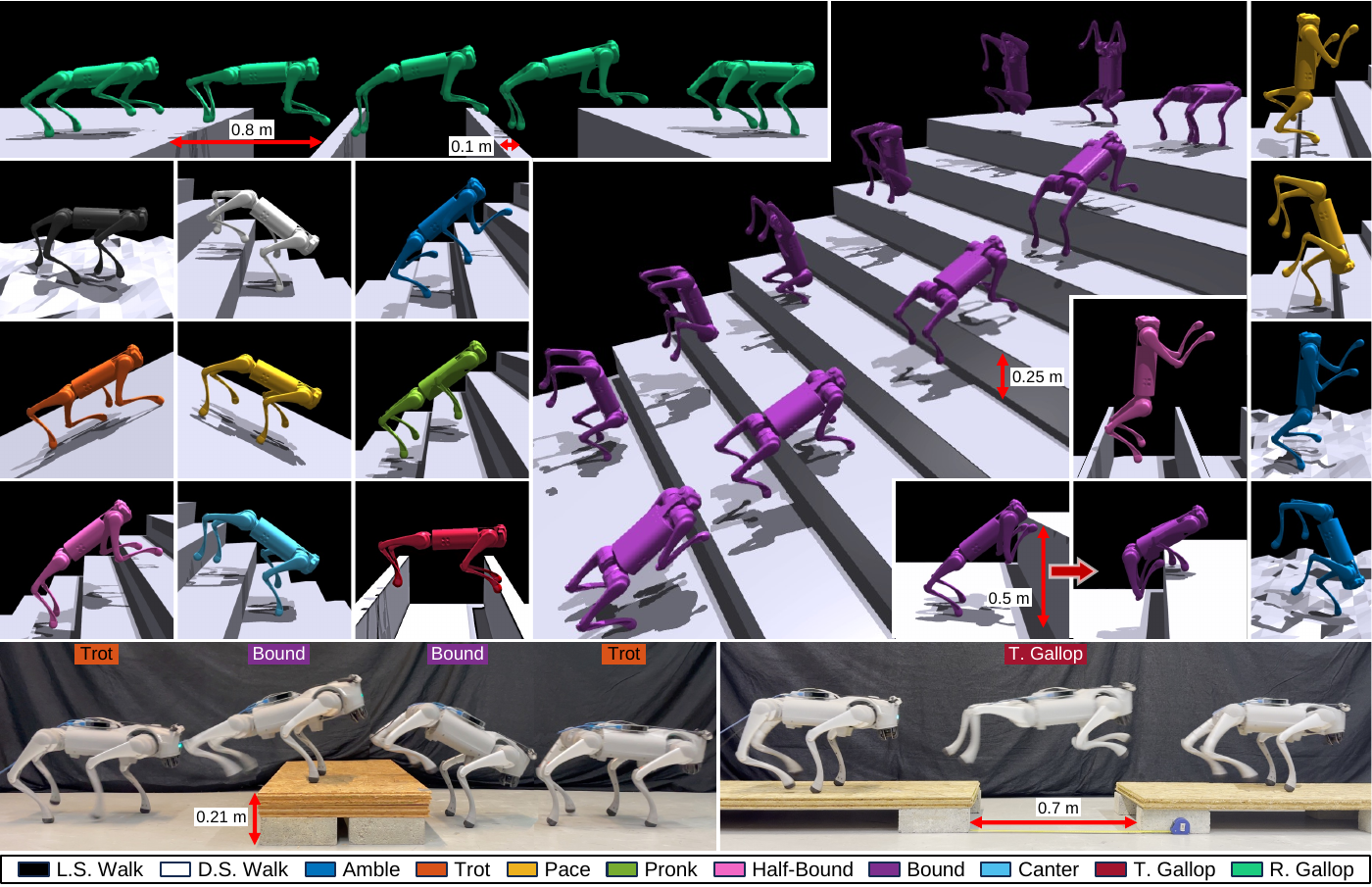}
    \caption{\textbf{Diverse gaits on challenging terrains.}
    Locomotion with 11 distinct gait couplings on slopes, stairs, gaps, rough terrain, and random blocks. Color indicates the gait:
    \textit{lateral sequence walk},
    \textcolor{walk2color}{diagonal sequence walk},
    \textcolor{amblecolor}{amble},
    \textcolor{trotcolor}{trot},
    \textcolor{pacecolor}{pace},
    \textcolor{pronkcolor}{pronk},
    \textcolor{halfboundcolor}{half-bound},
    \textcolor{boundcolor}{bound},
    \textcolor{cantercolor}{canter},
    \textcolor{tgallopcolor}{transverse gallop}, and
    \textcolor{rgallopcolor}{rotary gallop}. The same gait couplings are also used to learn bipedal-mode locomotion.
    \textbf{Bottom:} hardware deployment with the height map estimated onboard from a forward-facing depth camera.}
    \label{fig1}
\end{figure*}

\begin{figure*}[!t]
    \centering
    \includegraphics[width=\textwidth]{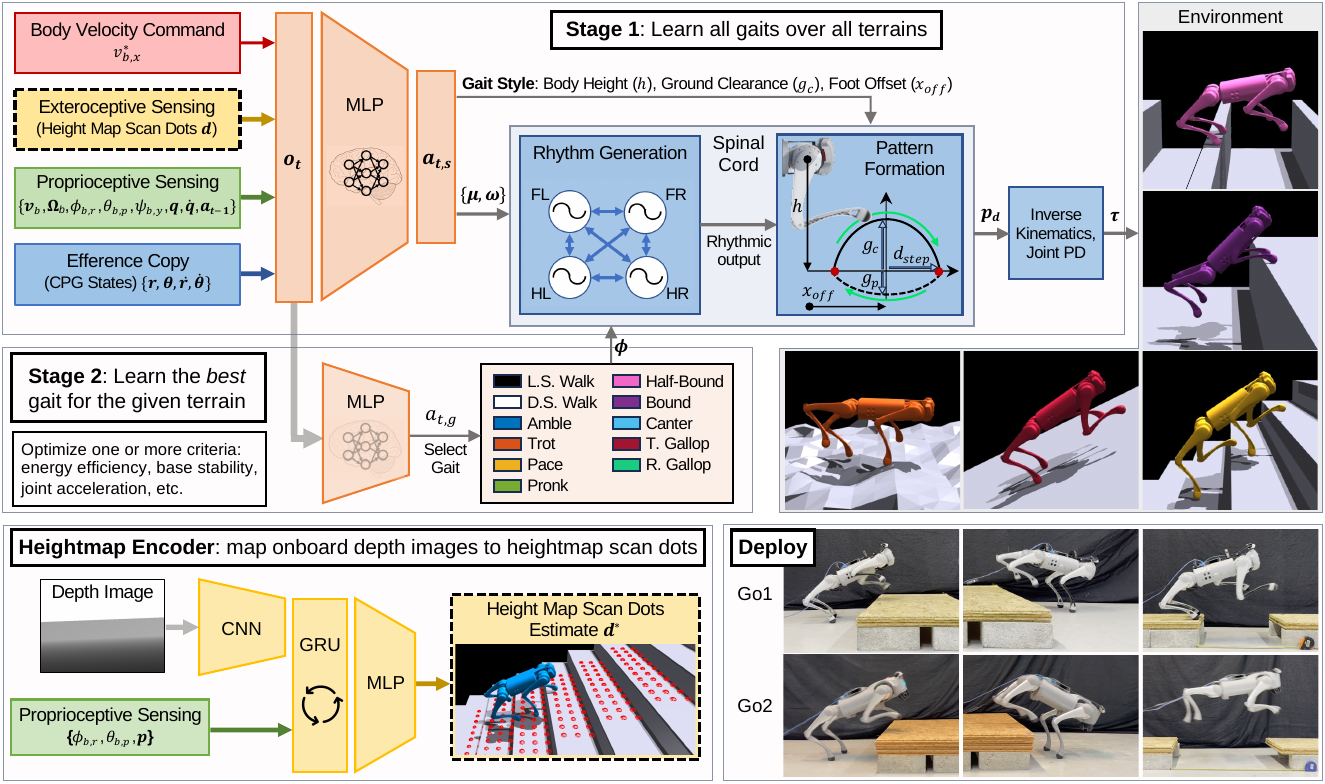}
    \caption{\textbf{Control architecture for terrain-adaptive gait learning and selection.}
    \textbf{Stage~1}: the policy observes a velocity command $v^{*}_{b,x}$, exteroceptive sensing (height-map scan dots $\bm{d}$), proprioceptive sensing (body velocities $\bm{v}_b$, $\bm{\Omega}_b$; orientation $\phi_{b,r}$, $\theta_{b,p}$, $\psi_{b,y}$; joint states $\bm{q}$, $\dot{\bm{q}}$; previous action $\bm{a}_{t-1}$), and CPG states ($\bm{r}$, $\bm{\theta}$, $\dot{\bm{r}}$, $\dot{\bm{\theta}}$), all as inputs to a multilayer perceptron (MLP). It outputs gait style parameters (body height $h$, ground clearance $g_c$, foot offset $x_{off}$) and CPG parameters ($\mu_i$, $\omega_i$) for each leg, coordinating the Rhythm Generation and Pattern Formation layers. Task-space foot positions $\bm{p}_d$ are converted to joint angles via inverse kinematics and tracked with PD control to produce torques $\bm{\tau}$. The enforced desired gait is provided as the coupling matrix $\bm{\Phi}$, resampled every few seconds during training.
    \textbf{Stage~2}: the Stage~1 network is frozen; a second policy receives the same observation as Stage~1 and selects among 11 gaits at each decision timestep, adapting gait selection online to changing terrain to optimize a user-defined criterion.
    \textbf{Heightmap encoder}: for hardware deployment, the privileged heightmap is replaced by an estimate produced onboard. An encoder predicts the height-map scan dots from depth images captured by a camera mounted on the robot. The images pass through a convolutional neural network (CNN), and the resulting embedding is concatenated with body orientation ($\phi_{b,r}$, $\theta_{b,p}$) and foot positions ($\bm{p}$) and passed to a single-layer gated recurrent unit (GRU), which lets the encoder integrate observations over time. A final MLP regresses the height-map scan dots, trained with DAgger. The Stage~1 and Stage~2 networks are unchanged. \textbf{Deploy}: with the encoder and camera onboard, the policies run on the physical Unitree Go1 and Go2.}
    \label{fig2}
\end{figure*}

Here we present exactly such a framework, with various snapshots shown in Fig.~\ref{fig1}. We combine DRL with networks of coupled oscillators representing a CPG\cite{ijspeert2008} to learn 11 distinct quadrupedal gaits, spanning the full Hildebrand taxonomy\cite{hildebrand1965symmetrical,hildebrand1977analysis} from lateral sequence walk to rotary gallop, across seven challenging terrain types. Training proceeds in two stages (Fig.~\ref{fig2}). In \textbf{Stage~1}, the gait coupling is randomly sampled and a single policy learns to coordinate gait parameters and style for every gait-terrain combination. The gait coupling is not learned but imposed through the coupling matrix of Eq.~\ref{eq:theta}, requiring the policy to learn how to locomote under each prescribed phase relationship while jointly adapting gait parameters and posture to the terrain. Crucially, this requires only a minimal reward function: we do not need any style rewards to enforce particular gaits, body heights, or foot clearances, unlike prior methods that require extensive reward engineering or expert data. The same velocity-tracking and regularization reward is used for all gaits and terrains. In \textbf{Stage~2}, the Stage~1 network is frozen and a lightweight second policy learns to \emph{select and transition between gaits} to optimize a user-specified criterion such as energy efficiency, base stability, or joint smoothness, using a separate reward function for each criterion. This decomposition is critical: when gait selection is optimized concurrently with low-level control, the policy typically converges to a single gait, a local optimum that fails to explore the full space of locomotion strategies. In contrast, by freezing the low-level controller and training a lightweight gait selector on top, gait selection can be re-optimized for different performance metrics by retraining Stage~2 alone, without retraining the underlying locomotion policy.

Our analysis reveals a striking result: \textbf{all 11 gaits appear as optimal under at least one terrain--metric--velocity combination}, reinforcing the biological intuition that gait diversity exists precisely because no single gait dominates universally. We further demonstrate strong generalization to diverse out-of-distribution conditions: novel gait couplings not seen during training, gait style overrides up to $2\times$ the training range maximum, and terrain difficulties almost 50\% beyond the training distribution. Furthermore, we demonstrate cross-embodiment transfer to larger quadrupeds with $5\times$ the training mass, deployment on two physical robot platforms, and bipedal-mode locomotion on challenging terrains by adding only a single pitch-tracking reward term.

\section*{Results}
\label{sec:results}

We develop a hierarchical bio-inspired architecture (Figure~\ref{fig2}) consisting of a policy trained with deep reinforcement learning (higher centers), a network of coupled oscillators mapped to task space foot trajectories (rhythm generator and pattern formation layers of the spinal cord), and sensory feedback from onboard sensors and internal CPG states (efference copy of the spinal cord). In the first stage of training, we learn all 11 gaits on all terrains, where the policy learns to modulate both the gait parameters (step frequencies and step lengths) as well as the locomotion style through the pattern formation parameters (i.e.~body height, swing foot ground clearance, foot offsets). This architecture allows the policy to learn to locomote with 11 quadrupedal gaits (Lateral Sequence Walk, Diagonal Sequence Walk, Amble, Trot, Pace, Pronk, Half-Bound, Bound, Canter, Transverse Gallop,  Rotary Gallop; refer to Supplementary Figure 1 for their definitions) and co-optimize the gait posture and step lengths and frequencies in conjunction with the terrain. In the second stage of training, we optimize for different performance criteria in order to determine which of the discrete gaits to use on which terrain. With this two-stage process, we ensure that all gaits are explored on all terrains in Stage~1, before optimizing gait selections and transitions in Stage~2 based on specific criteria. For hardware deployment, a third network estimates the height map onboard from depth images.

\begin{figure*}[!t]
    \centering
    \includegraphics[width=\textwidth]{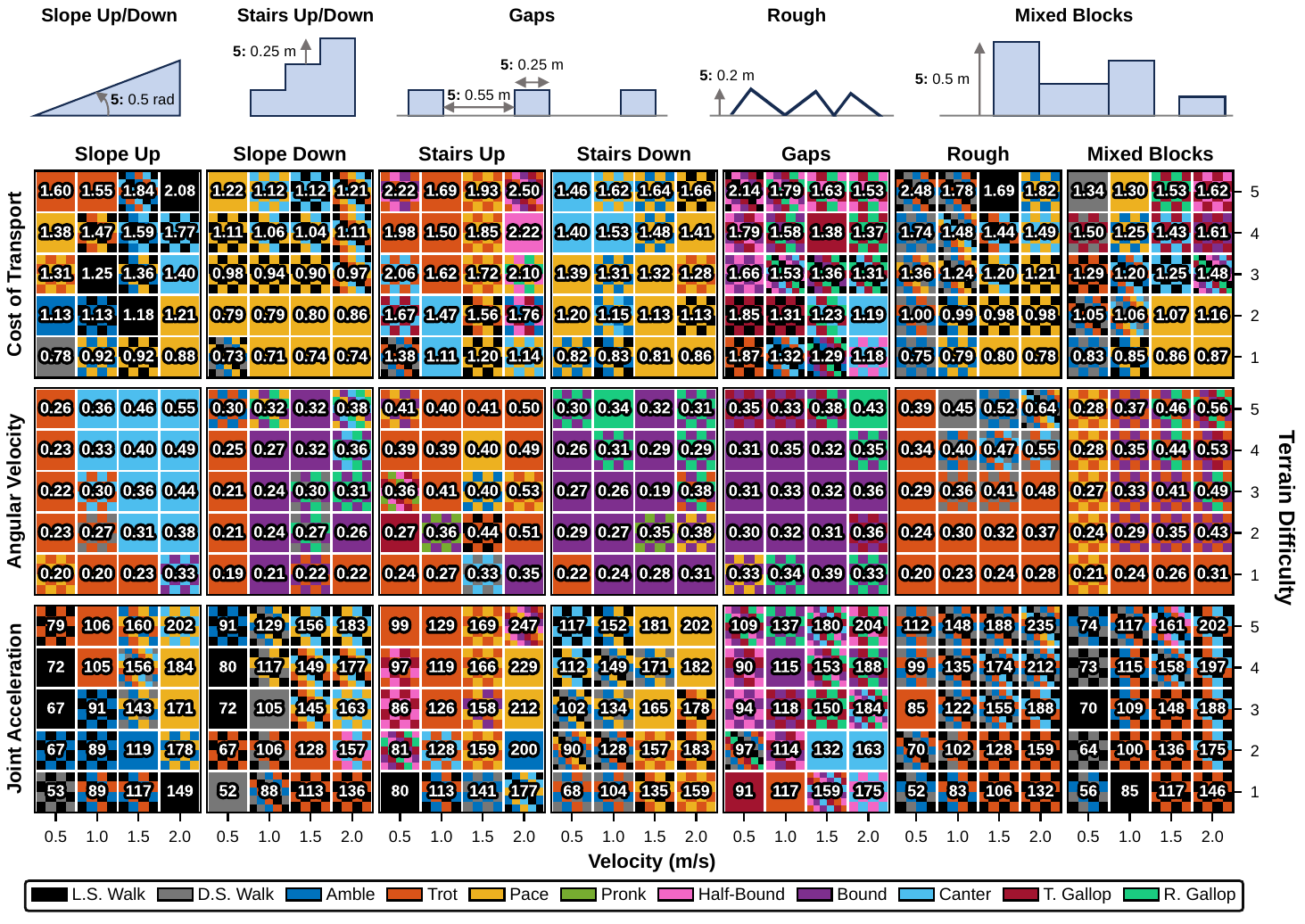}\\
    \vspace{-0.5em}
    \caption{\textbf{Optimal gaits across terrains and metrics.}
    \textbf{Top}: test terrains ranked in difficulty from 1 (easiest) to 5 (most difficult). Difficulty~5 corresponds to the steepest slopes, tallest steps, widest gaps (and smallest platforms), most uneven rough terrain, and largest random mixed blocks.
    \textbf{Bottom}: the gait achieving the lowest mean Cost of Transport, base angular velocity, and joint acceleration across five terrain difficulties and four velocity commands (0.5--2\,m/s), averaged over 100 parallel agents on 10\,m terrains. Lower values are better for each metric, and the corresponding gait is shown for each terrain difficulty and velocity combination. When multiple gaits achieve metric values within 5\% of the minimum, their colors appear in a checkerboard pattern. Remarkably, all 11 gaits appear as optimal under at least one scenario.}
    \label{fig3}
\end{figure*}

\subsection*{Learning all gaits over all terrains}

We first analyze how the agent learns to coordinate gait style and gait parameters to locomote with enforced gaits across diverse terrains. We create seven test terrains: slopes up/down, stairs up/down, gaps, uniform random rough terrain, and random mixed blocks. Each terrain is scaled in difficulty from level~1 to~5, with 5 being the most challenging (Fig.~\ref{fig3}, top). For each terrain, we command 100 robots in parallel to traverse 10\,m courses at four velocities (0.5, 1.0, 1.5, and 2.0\,m/s) and analyze the metrics listed in Methods. We present the gaits that minimize mean cost of transport (COT), mean base angular velocity, and mean joint acceleration in Fig.~\ref{fig3}, and additional metrics are analyzed in Supplementary Figs.~3--11. For these three metrics, lower values indicate better energy efficiency, better base stability, and smoother joint motion, respectively. We consider `ties' between gaits as values within 5\% of the minimum, leading to checkerboard patterns. Throughout, we use the term \emph{optimal gait} for a stated metric to mean the gait that minimizes that metric among the 11 gaits, for a given terrain, difficulty, and velocity. 

Remarkably, all 11 gaits can be found to minimize one of these metrics during some particular terrain scenario, highlighting the value of being able to generate multiple gaits and select the most suitable one for a given terrain, metric, and velocity. Notably, it is rarely the case that a single gait is best under all three metrics for a given terrain. However, there are important trends. On relatively flat terrain (difficulty~1), similar to in nature, gaits such as the walks, amble, and pace are generally most energy-efficient. As can be expected, the trot gait generally keeps a flat base and minimizes base angular velocity across more than half of the evaluated velocity--terrain combinations, and also generally has low joint acceleration. This gait also typically emerges when training DRL controllers end-to-end for relatively flat terrains\cite{ji2022concurrent,margolis2022rapid,miki2022learning,rudin2022anymalisaac}. However, from an energy-efficiency perspective, the trot is only best in roughly 25\% of evaluated velocity--terrain combinations. When traversing difficult terrains with wide gaps, gaits with approximate fore--hind symmetry like half-bound, bound, transverse gallop, and rotary gallop minimize energy and joint acceleration, and have lower angular velocity. The bound gait in particular typically has minimal roll and yaw rates due to symmetry. Somewhat surprisingly, even the pronk gait appears as optimal for one of these criteria: minimizing mean angular velocity in a few instances on stairs, which can be explained by the enforced symmetry synchronizing well with a particular set of stairs.

We also compare against a baseline policy trained with the same terrain curriculum and reward function, but \emph{without} inter-oscillator coupling, reducing the phase dynamics to $\dot{\theta}_i = \omega_i$. This baseline can freely coordinate each limb to form its own gaits. Although we might expect it to adapt more efficiently since it is not constrained to a particular gait, we find it is not optimal in most cases (Supplementary Figs.~3--11). This highlights that choosing the right gait is critical to optimizing performance, and further shows that gaits emerging during unconstrained learning can stay stuck in local optima. The baseline consistently locomotes with larger steps and lower step frequencies across all terrains, and has the lowest joint acceleration in most velocity--terrain scenarios, suggesting it may have optimized this quantity over other metrics like angular velocity or COT. More broadly, Supplementary Table~3 situates the framework relative to existing terrain-aware controllers\cite{cheng2024extreme,zhuang2023robot,hoeller2024anymal,yang2024agile}, which demonstrate challenging obstacle traversal but do not provide the same combination of a commandable gait repertoire, systematic gait-by-gait comparisons, and velocity tracking across commanded speeds.

\subsection*{Optimizing gait selection for specific criteria}

The previous section showed that even when considering only one metric, it is not always obvious which gait to select under a particular terrain scenario due to ties. Additionally, it is infeasible to roll out the policy on every possible terrain with every possible velocity. Therefore, we train a second neural network to select the gaits to be used by the Stage~1 policy, to optimize for particular criteria. This second training stage also importantly learns \emph{when} to transition between gaits to continue optimizing the chosen criterion. The selector receives the same observations as Stage~1, including terrain heightmaps and commanded velocity, and selects one of the 11 gaits at each control timestep.

\begin{figure*}[!t]
    \centering
    \includegraphics[width=\textwidth]{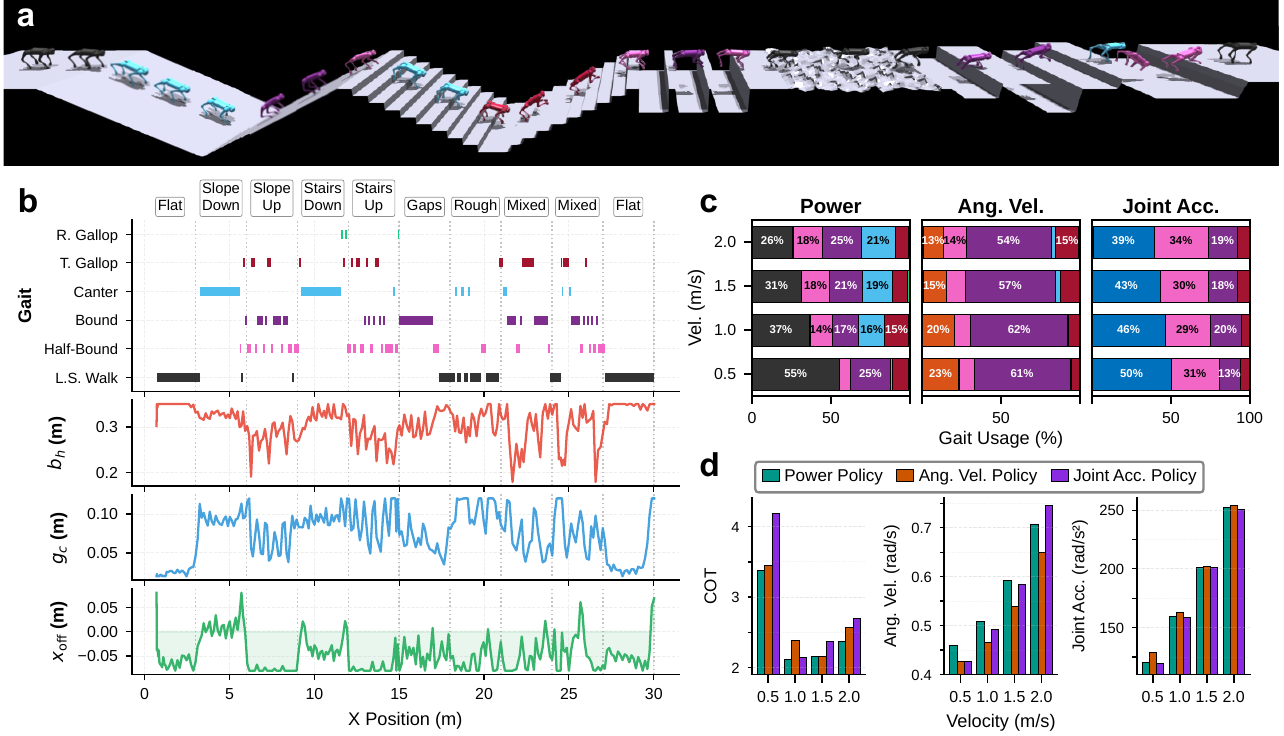}\\
    \vspace{-0.3em}
    \caption{\textbf{Stage~2 gait selection on a parkour course.}
    \textbf{a}: parkour course incorporating all considered terrain types. Robot snapshots show results from the power-minimizing policy.
    \textbf{b}: gaits selected by the power-minimizing Stage~2 policy at 1.5 m/s along the course shown in \textbf{a}, together with the corresponding gait style parameters (body height, foot ground clearance, foot offset) output by the Stage~1 policy. A single gait is active at each timestep, plotted against course position. The Stage~2 policy sometimes transiently selects a gait (rotary gallop), or switches rapidly between similar gaits (bound and half-bound, or bound and transverse gallop), which can occur when two gaits are close to tied on the optimized criterion (Fig.~\ref{fig3}). The Stage~1 policy modulates the gait style parameters intuitively, e.g.~shifting the foot offset forward when descending slopes or stairs. 
    \textbf{c,d}: comparing gait selections and metrics on the parkour course for three Stage~2 policies.
    \textbf{c}: percentage of time each gait is selected by policies trained to minimize power, angular velocity, and joint acceleration, respectively, across four commanded velocities. Each regularly selects a unique gait not used by the others.
    \textbf{d}: mean metric values for each policy, computed over the entire course. Each policy outperforms the others on its respective targeted metric.
    }
    \label{fig4}
    \vspace{-1em}
\end{figure*}

We focus our discussion here on three separately trained Stage~2 policies to individually minimize (1)~power, (2)~mean base angular velocity, and (3)~mean joint acceleration, on the same terrains from Stage~1. Each policy uses a different optimization reward (Supplementary Fig.~12), while the Stage~1 controller remains frozen and continues to adapt the gait parameters and locomotion style. For testing, we design a parkour scenario incorporating each of the considered terrains, as depicted in Fig.~\ref{fig4}-a. The robot snapshots show the gaits selected by the policy that optimized for power consumption while locomoting at 1.5 m/s. 
Fig.~\ref{fig4}-b shows the complete sequence of gait selections from the same 1.5 m/s rollout of the power-minimizing Stage~2 policy, together with the corresponding base height, ground clearance, and foot offsets output by the Stage~1 policy.
We see that the Stage~1 policy has learned to modulate these parameters extensively and intuitively, for example shifting the foot offset forward when going down a slope or stairs. Qualitatively, the Stage~2 policies have learned several important gait transitions based on velocity and terrain: for example, when walking or trotting at low velocity and approaching a gap or large step, they switch to a bound-style gait to jump over the obstacle.

We compare the three Stage~2 policies on the test parkour course across four velocities in Fig.~\ref{fig4}c-d, which shows the percentage of time each gait is selected (Fig.~\ref{fig4}-c) and the resulting mean metric values over the entire course (Fig.~\ref{fig4}-d). The three policies each have a unique gait not used by the others, mainly for relatively flat terrain locomotion: lateral sequence walk for minimizing power, trot for minimizing base angular velocity, and amble for minimizing joint acceleration. These results are consistent with the evaluation of the Stage~1 policy. Notably, as the commanded velocity increases, the percentage of time used for these lower-velocity gaits decreases, with the more aggressive gaits like (half-)bound being used a larger fraction of the time. We verify that the desired metrics are actually minimized: each policy outperforms the others on its targeted metric (Fig.~\ref{fig4}-d). Additional Stage~2 policies trained with alternative reward functions, including those that \emph{maximize} undesirable quantities to reveal which gaits are `worst', are reported in Supplementary Fig.~12.

\subsection*{Generalization}

We test the generalizability of our framework to tasks outside those seen during training.

\begin{figure*}[!t]
    \centering
    \includegraphics[width=\textwidth]{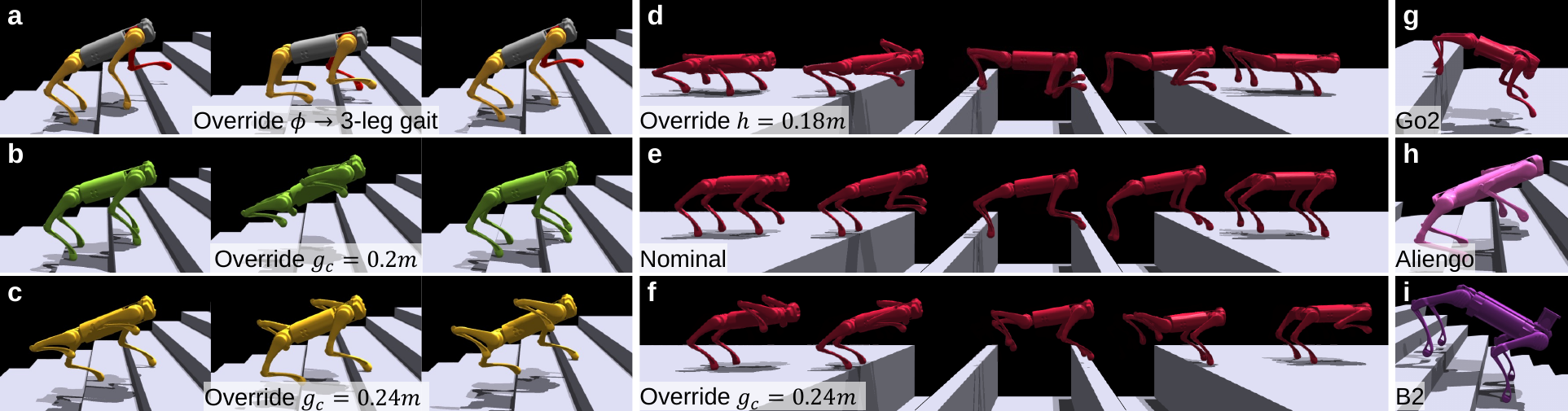}
    \caption{\textbf{Generalization to out-of-distribution conditions.}
    \textbf{a}: zero-shot locomotion with a novel 3-legged gait coupling not seen during training.
    \textbf{b}: pronk up stairs with overridden ground clearance ($g_c = 0.2$\,m, approximately $1.7\times$ the training maximum).
    \textbf{c}: pace up stairs with ground clearance overridden to $2\times$ the training maximum ($g_c = 0.24$\,m).
    \textbf{d--f}: gap traversal with different gait style parameter settings. \textbf{d}: base height forced to the training minimum (0.18\,m). \textbf{e}: nominal traversal without overrides. \textbf{f}: ground clearance forced to $2\times$ the training range maximum (0.24\,m).
    \textbf{g--i}: generalization to other quadruped platforms. \textbf{g}: Unitree Go2 (roughly the same size as Go1), \textbf{h}: Unitree Aliengo ($2\times$ mass, $1.5\times$ height), \textbf{i}: Unitree B2 ($5\times$ mass, $2.15\times$ height).}
    \label{fig5}
\end{figure*}

\textbf{Novel gaits.} While in this work we focused on quadruped gaits found in nature, there are infinitely many sets of phase couplings that could be imposed between different oscillators. We test one such combination with three legs in phase and a fourth out of phase by $\pi$, forming a sort of limping gait coupling (Supplementary Figure 1). While the agent never saw such a coupling during training, we find it is still able to adapt zero-shot to this new phase coupling, even in a difficult scenario of locomotion up stairs at a high velocity (Fig.~\ref{fig5}a).

\textbf{Override gait parameters.} During training the agent modulates the base height within 0.18--0.35\,m and the swing foot ground clearance within 0.02--0.12\,m, corresponding to 47--91\% and 5--31\% of the maximum leg extension. The agent has optimized how to modulate these parameters together with the foot offsets, amplitudes, and frequencies to coordinate locomotion over varied terrains. We test overriding these parameters away from their optimized values and find that the agent can still locomote with parameters never seen during training, for example with a foot ground clearance up to twice the training maximum (0.24\,m), as shown in Fig.~\ref{fig5}b,c. It is possible to override these parameters even on challenging terrain at fast velocities over stairs or wide gaps. We observe the effects of overriding parameters on a gap traversal task (Fig.~\ref{fig5}d--f): forcing the base height to its minimum, and forcing the ground clearance to $2\times$ the training maximum, both of which make the task substantially harder. In both cases, the agent still successfully traverses the terrain. 

\textbf{Out-of-distribution terrains.} In Supplementary Video 1, we show generalizability to several scenarios that are much more challenging than those seen during training. For example, the maximum gap width during training is 0.55\,m, yet at test time the robot can gallop across gaps that are almost 50\% wider (0.8\,m), demonstrating robust extrapolation beyond the training distribution. Additionally, the robot can bound over a 0.55\,m wall (10\% higher than in training), or gallop across 0.4\,m stairs (60\% higher than seen in training). 

\textbf{Generalization to other quadruped platforms.} We test whether the policy trained on the Unitree Go1 can transfer zero-shot to other quadruped robots with substantially different body scales~\cite{shafiee2024manyquadrupeds}. Without any retraining, the same policy successfully controls the Unitree Go2 (roughly the same size as the Go1), the Unitree Aliengo ($2\times$ mass, $1.5\times$ height), and the Unitree B2 ($5\times$ mass, $2.15\times$ height) across challenging terrains (Fig.~\ref{fig5}g--i). This cross-embodiment transfer is enabled by the CPG-based action space: because the policy outputs abstract oscillator parameters and gait style commands rather than direct joint torques, the mapping to physical joint angles is handled by per-robot inverse kinematics, allowing the learned locomotion strategies to generalize across body scales. Any quadrupedal platform can also be trained directly.

\subsection*{Hardware deployment}

Both policies transfer to physical robots. An encoder maps onboard depth images to height-map estimates (Fig.~\ref{fig2}), leaving the Stage~1 and Stage~2 networks unchanged, and the same policies drive either the Unitree Go1 or the Go2. Stage~1 executes any commanded gait, including a bound or gallop over a 0.21\,m obstacle and across a 0.7\,m gap at 1\,m/s, a gap wider than any presented during training. Stage~2 selects and transitions between gaits online from the onboard estimate, walking or trotting on flat ground and switching to a bound or gallop at obstacles and gaps (Supplementary Video~2).

\begin{figure*}[!t]
    \centering
    \includegraphics[width=\textwidth]{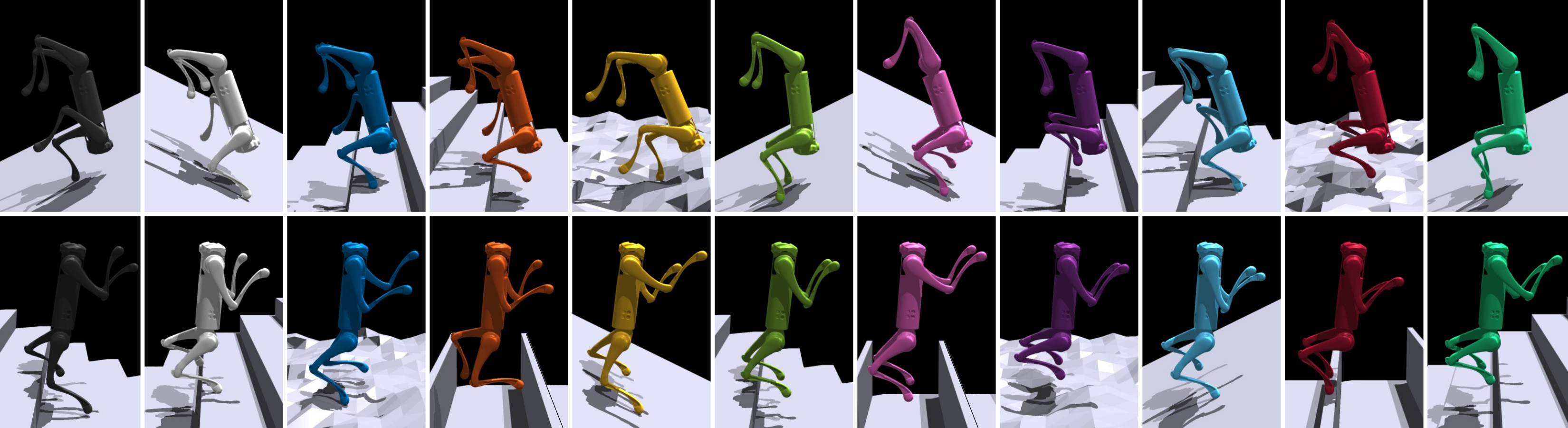}
    \caption{\textbf{Bipedal-mode locomotion on diverse terrains with all gait couplings.} The robot locomotes on the front legs (handstand, positive pitch) and on the rear legs (negative pitch), using all 11 quadrupedal gait couplings in each configuration. Only one additional reward term (for base pitch) is needed, without any expert data or reference motions. The robot autonomously transitions from quadrupedal stance to a front-leg handstand, while rear-leg locomotion begins from an upright stance.
    }
    \label{fig6}
\end{figure*}

\subsection*{Bipedal-mode locomotion}

Our framework generalizes beyond the classic quadrupedal gaits. By adding a single reward term to the Stage~1 training procedure to track the robot pitch, we can learn bipedal-mode locomotion with all 11 quadrupedal gait couplings, on either the front legs or the rear legs (positive or negative pitch, respectively). Prior works demonstrating quadruped handstands typically need significant reward function tuning\cite{kim2025stage,xiao2025learning,zakka2025mujocoplayground} and/or expert reference data and specialized training curricula\cite{fuchioka2023optmimic,smith2023learning,tu2025continuous}. Additionally, previous handstand locomotion has largely been demonstrated on flat or uneven terrain\cite{xiao2025learning}, with stair traversal limited to descent\cite{cheng2024extreme}. Here, we show we can use any quadruped leg coupling for bipedal-mode locomotion by adding only a single pitch term, allowing robust control on diverse and difficult terrains. For front-leg handstands, no special initialization is required, as the robot autonomously transitions from quadrupedal to bipedal locomotion. For rear-leg locomotion, training begins with the robot standing upright on its rear legs. We note the emergence of step skipping when commanded to traverse the steps at high velocities, both up and down stairs. Snapshots of all gait couplings being used in biped mode on challenging terrains are shown in Fig.~\ref{fig6}.

\section*{Discussion}
\label{sec:discussion}

We have presented a framework for optimizing quadruped gaits as a function of terrain, velocity, and user-defined metrics. By combining bio-inspired CPG oscillators with deep reinforcement learning in a two-stage architecture, we can execute, evaluate, and select among 11 distinct quadrupedal gaits on challenging terrains---a capability not demonstrated by prior learning-based or model-based approaches, to the best of our knowledge.

A central finding is that every gait is optimal under some terrain--metric--velocity scenario. This is not what one would predict. The natural expectation could have been that a few gaits dominate and the remainder persist for evolutionary reasons, particularly the pronk, which is symmetric, expensive, and no obvious candidate for optimality under any criterion. That every gait in the taxonomy wins somewhere suggests that this diversity has functional significance beyond the familiar observation that animals change gait with speed. Individual assignments are also often unexpected: the trot, which most learned controllers converge to, is most energy-efficient in only about a quarter of the evaluated velocity--terrain combinations. This result has important implications for both biology and robotics. In biology, it is consistent with the hypothesis that the diversity of quadrupedal gaits evolved, at least in part, because no single locomotion strategy dominates across the range of terrains and performance demands animals encounter\cite{alexander1984gaits}. In robotics, it challenges the common practice of either letting a single gait emerge from training, or fixing one gait for all scenarios. Our results suggest that principled gait selection can yield measurable improvements in energy efficiency, stability, and robustness.

These assignments are predictions rather than confirmations of known biology. Much of the experimental evidence for gait selection comes from flat-terrain energetics\cite{hoyt1981gait}, where the mechanism is itself debated: in horses the trot--gallop transition occurs below the energetically optimal speed and appears triggered by musculoskeletal force rather than metabolic cost\cite{farley1991mechanical}. Studies of slope locomotion have largely examined changes in posture and limb mechanics within a walking gait, rather than comparing alternative gait choices\cite{carlsonkuhta1998upslope,smith1998downslope}. Muscle-driven horse models can reproduce several gait patterns and some spontaneous transitions, but have not recovered spontaneous transitions to asymmetric gaits such as galloping\cite{vanbijlert2024muscle}. Each entry in Fig.~\ref{fig3} is therefore a testable prediction about gait performance under a specified terrain, velocity, and criterion.

The two-stage architecture is essential. When gait selection is jointly optimized with low-level control in a single training phase, the policy typically converges to a single gait, leaving alternative locomotion strategies largely unexplored. By first ensuring all gaits are explored on all terrains (Stage~1), then training a lightweight gait selector (Stage~2), we achieve both broad coverage and targeted optimization with minimal reward engineering. Because Stage~1 is frozen and competent across the whole gait repertoire, the selection criterion becomes a free parameter of the method rather than a property fixed at design time. Optimizing a new criterion requires training only a lightweight Stage~2 selector, which takes minutes, rather than retraining the Stage~1 locomotion policy, which takes hours to achieve near-perfect traversal success across all gaits on the most challenging terrains. This makes the range of criteria in Fig.~\ref{fig4} and Supplementary Fig.~12 tractable. Because the criterion is explicit, the resulting gait preferences can be interpreted through the performance trade-offs in Fig.~\ref{fig3}, while Fig.~\ref{fig4} shows where transitions occur along a course. The uncoupled baseline comparison further validates this design: despite having no gait constraint, the baseline is outperformed by the best prescribed gait in most scenarios, demonstrating that unconstrained learning frequently produces suboptimal gait choices. 

The role of inter-oscillator coupling strength deserves particular attention. We find that sufficiently strong coupling is necessary to enforce the desired gait pattern; when coupling is weak, the policy can effectively override the  desired gait. We achieve high gait-tracking accuracy across all 11 gaits, with typically 95--100\% of commanded stance time realized as ground contact on almost all terrains and velocities (Supplementary Fig.~2). The main reductions occur over wider gaps, where suitable footholds are unavailable: coupling constrains oscillator phases rather than physical ground contacts. For example, over the widest gaps, a commanded pronk cannot place all four feet on the far platform simultaneously, and the fore and hind pairs touch down separately. The trade-off between enforcing gait patterns and allowing flexibility mirrors a design tension present in earlier CPG-RL work, where coupling was sometimes omitted to grant the policy more freedom\cite{bellegarda2022cpgrl} and later reintroduced to improve structure and transferability\cite{bellegarda2024visualcpgrl}. Interestingly, a related pattern can also be observed in biological locomotion. In \emph{Drosophila}, inter-leg coordination shifts from weak or absent coupling at low walking speeds to strong coupling at high speeds, with sensory feedback also contributing to coordination\cite{berendes2016speed}. This suggests that biological CPGs may not operate with fixed coupling strengths, but rather modulate them as a function of speed and context---possibly through descending commands from higher brain centers. Our framework, in which coupling strength is a tunable parameter, could serve as a computational testbed for investigating such hypotheses.

The framework's generalization capabilities are notable. Zero-shot adaptations to novel gait couplings, extreme parameter overrides (swing-foot ground-clearance overrides up to twice the training maximum), out-of-distribution terrains (almost 50\% wider gaps), and cross-embodiment transfers to a robot with $5\times$ the training mass, together with deployment on two hardware platforms, suggest that our architecture learns a robust relationship between oscillator dynamics and locomotion, rather than memorizing specific gait--terrain pairs. Finally, the ability to achieve bipedal-mode locomotion on challenging terrains by adding only a single pitch-tracking reward term, without expert data or reference motions, further demonstrates the versatility of the coupled-oscillator representation.

Several directions remain open. Because Stage~1 remains frozen, new performance criteria in Stage~2 are optimized through gait selection rather than directly re-optimizing the underlying gait parameters and locomotion styles; adapting these while preserving the learned gait repertoire is a natural direction for future work. In simulation, failures concentrate on mixed blocks and arise when an obstacle is met off-center, since our locomotion controller acts primarily in the sagittal plane. Adding lateral foot placement control\cite{bellegarda2022cpgrl,bellegarda2024visualcpgrl} could improve recovery in these cases and extend the framework to terrains requiring sharp turns. The seven evaluation terrains were chosen to span a range of challenging terrain geometries. However, the framework itself is not specific to them, and the same procedure could in principle apply to other terrains we have not addressed here, such as deformable substrates, mud, snow, and water. While we demonstrate cross-embodiment transfer across body scales within the Unitree family, extending to morphologies with fundamentally different limb kinematics or musculoskeletal actuation would test whether the CPG abstraction generalizes beyond rigid-body torque-controlled platforms, and would strengthen the biological relevance of the framework's predictions. An intriguing direction would be to make inter-oscillator coupling strength itself a learnable or speed-dependent variable, allowing the framework to discover whether the biological pattern of speed-modulated coupling\cite{berendes2016speed} emerges as an optimal control strategy. Finally, integrating gait selection with higher-level planning would enable long-horizon navigation in which terrain, energy consumption, and task constraints jointly determine not only the path but the locomotion strategy along it. More broadly, our framework opens avenues for computational studies of biological gait selection by providing a testbed in which terrain, morphology, and performance criteria can be systematically varied.

\section*{Methods}
\label{sec:methods}

\subsection*{Rhythm generation and pattern formation}

We use a network of abstract oscillators to enforce and coordinate the different gait patterns, specifying the desired contact sequence through interlimb phase relationships. Following CPG-RL\cite{bellegarda2022cpgrl,bellegarda2025allgaits}, the Rhythm Generation layer uses abstract oscillators of the form:
\begin{align}
\ddot{r}_i &= a \left(\frac{a}{4} \left(\mu_i - r_i \right) - \dot{r}_i \right) \label{eq:r} \\
\dot{\theta}_i &= \omega_i + \sum_{j} r_j w_{ij} \sin(\theta_j - \theta_i - \phi_{ij}) \label{eq:theta}
\end{align}
where $r_i$ and $\theta_i$ are the amplitude and phase of oscillator~$i$, $\mu_i$ and $\omega_i$ are the intrinsic amplitude and frequency, $a$ is a positive convergence constant, and the couplings are defined by weights $w_{ij}$ and phase biases $\phi_{ij}$. For a quadruped with one oscillator per leg, the $4 \times 4$ coupling matrices $\bm{\Phi}$ define the phase offsets determining each gait's contact sequence. The timings for the 11 gaits are visualized in Supplementary Fig.~1. As a brief example, in the pace gait, a 0.5 phase offset represents a $\pi$ offset between the left and right limbs. With appropriately strong coupling weights (i.e.~$w_{ij}=10$), these coupling matrices enforce the desired oscillator phase relationships. The gait thus acts as a constraint on the phase dynamics, and is not a quantity the Stage~1 policy can optimize directly.

In the Pattern Formation layer, the oscillator states are mapped to foot trajectories, and corresponding joint angles are computed with inverse kinematics. The desired foot position coordinates are:
\begin{align}
x_{i,\text{foot}} &= x_{off} - d_{step} (r_i - 1) \cos(\theta_i) \\
z_{i,\text{foot}} &= \begin{cases}
    -h + g_c\sin(\theta_i) & \text{if } \sin(\theta_i) > 0 \\
    -h + g_p\sin(\theta_i) & \text{otherwise}
\end{cases}
\end{align}
where $d_{step}$ is the maximum step length, $x_{off}$ is the foot offset relative to the hip, $h$ is the robot height, $g_c$ is the maximum swing foot ground clearance, and $g_p$ is the maximum ground penetration during stance. Together with the CPG parameters, the policy co-optimizes $h \in [0.18, 0.35]$\,m, $x_{off} \in [-0.08, 0.08]$\,m and $g_c \in [0.02, 0.12]$\,m to adapt its locomotion capabilities depending on the terrain. $g_p$ is randomly sampled within $[0, 0.015]$\,m. These style parameters enter the pattern formation layer analytically and may therefore be set outside their trained ranges at test time (Fig.~\ref{fig5}b--f).

\subsection*{Stage~1: learning all gaits}

In the first stage of training, we train a single policy, where gaits and gait-transitions are randomly sampled from among the 11 possibilities to ensure that the agent learns to coordinate each of the different gaits on any terrain. While we cannot expect each gait to perform well on a given terrain, this ensures the agent sufficiently explores each contact sequence, in order to not overfit to one particular gait. The random transitions between different gaits ensure that a future gait selector network can switch between gaits without the robot falling.

\textbf{Observation space.}
The observation includes a forward velocity command $v^{*}_{b,x}$, exteroceptive sensing (a $17 \times 11$ height-map grid at 0.1\,m intervals around the robot base), proprioceptive sensing (body linear ($\bm{v}_b$) and angular ($\bm{\Omega}_b$) velocities, orientation ($\phi_{b,r}$, $\theta_{b,p}$, $\psi_{b,y}$), joint positions ($\bm{q}$) and velocities ($\dot{\bm{q}}$)), the last action chosen by the policy, internal CPG states ($\bm{r}$, $\dot{\bm{r}}$, $\bm{\theta}$, $\dot{\bm{\theta}}$), and the current gait index. The CPG states represent an efference copy of the spinal cord\cite{von1950reafferenzprinzip,crapse2008corollary} and serve as noise-free gait state information. 

\textbf{Action space.}
The agent selects gait posture parameters (body height $h$, ground clearance $g_c$, foot offset $x_{off}$) and modulates intrinsic oscillator parameters ($\mu_i$, $\omega_i$) for each leg $i$, yielding $\bm{a}_{t,s} = [h, g_c, x_{off}, \bm{\mu}, \bm{\omega}] \in \mathbb{R}^{11}$. Actions are selected at 33\,Hz with $\mu_i \in [1, 2]$ and $\omega_i \in [0, 45]$\,rad/s. The strong coupling enforces the relative offsets between different oscillators, meaning the agent is forced to learn parameters to locomote with the particular gait that is currently selected.

\textbf{Reward function.}
We retain the minimal reward structure and weights from our previous 
work\cite{bellegarda2022cpgrl,bellegarda2025allgaits}, without tuning them specifically for the gait comparisons in this study. The reward function primarily tracks commanded forward velocity, while penalizing undesired lateral and vertical body velocities, base angular velocities, mechanical power, and changes in policy actions. Additional regularization discourages actions outside the clipping range and deviations from the intended straight-line trajectory. The terms and weights are summarized in Supplementary Table~2. Some of these terms overlap with, but do not exactly match, the three primary evaluation metrics: mechanical power is related to cost of transport, the base angular velocity penalty to the mean angular velocity metric, and changes in policy actions only indirectly to joint acceleration. The resulting gait rankings thus depend on the common training objective and its relative weights. Unlike other reinforcement learning approaches, we do not require any `style' rewards to enforce particular gaits, body heights, or foot ground clearances, which the policy directly controls through the CPG interface. The same reward function is used for all 11 gaits and all terrains.

For bipedal-mode locomotion, we add a pitch-tracking reward $r_{\mathrm{pitch}} = \exp(-(\theta_{b,p}^{*}-\theta_{b,p})^2/0.25)$, with weight 5 and target pitches of $+\pi/2$ and $-\pi/2$ for front- and rear-leg locomotion, respectively. The existing velocity-tracking and stabilization terms are adjusted to the corresponding body axes in the upright configuration. For rear-leg locomotion, training begins from an upright pose.

\subsection*{Stage~2: learning gait selection}

After Stage~1 training is completed, the agent is capable of locomoting with all 11 gaits over each of the various terrains. With the Stage~1 network frozen, we train a new neural network in Stage~2 which takes the same observations and outputs an 11-dimensional gait-selection vector $\bm{a}_{t,g} \in \mathbb{R}^{11}$. The selected gait is $g_t = \arg\max_{k}(\bm{a}_{t,g})_k$, which determines the corresponding coupling matrix $\bm{\Phi}$ (Supplementary Fig.~1). We train separate Stage~2 policies to optimize different performance criteria individually, as shown in Supplementary Figure 12. Each gait selector policy uses a criterion-specific reward as its optimization objective, together with an action-rate penalty to discourage frequent gait switching. The frozen Stage~1 controller continues to track the commanded velocity and adapt gait parameters and locomotion style, while Stage~2 optimizes only gait selection. This allows both selecting and transitioning between gaits as a function of the terrain and commanded velocity to optimize the chosen performance criterion.

\subsection*{Heightmap encoder for onboard terrain perception}

The controllers of Stages~1 and~2 receive an exteroceptive heightmap that, in simulation, is sampled directly from the terrain. For onboard deployment we replace this privileged input with an estimate produced from a single forward-facing Intel RealSense D435i camera, so that no external motion capture or pre-built terrain map is required at test time. The locomotion and selection networks remain frozen, and only the encoder is trained.

\textbf{Architecture.} Following the depth-to-height-map distillation of Yang et al.\cite{yang2024agile}, a recurrent encoder maps the depth stream and proprioception to the height-map. Each depth image ($48\times16$, normalized to $[0,1]$) is embedded by a small convolutional network. This embedding is concatenated with the body orientation ($\phi_{b,r}$, $\theta_{b,p}$) and the foot positions ($\bm{p}$) expressed in a gravity-aligned base frame, and passed to a single-layer gated recurrent unit (GRU) with 256 hidden units. The recurrence lets the encoder integrate observations over time, allowing it to retain information about terrain that has passed beneath the robot or is no longer within the camera's field of view. A three-layer multilayer perceptron ($256$ units per layer) then regresses the height-map used by the Stage 1 and 2 controllers. At deployment this predicted height-map is substituted for the privileged terrain input, closing the perception--action loop entirely onboard. 

\subsection*{Training details}

We use Isaac Gym with PhysX\cite{makoviychuk2021isaac,rudin2022anymalisaac} and the Unitree Go1 quadruped. We simulate 4,096 robots in parallel on a single NVIDIA RTX 3090 GPU. Stage~1 training takes a few hours to fully converge on all terrains with Proximal Policy Optimization (PPO)\cite{schulman2017proximal}; Stage~2 trains on the order of minutes as the task is much simpler. The Stage~1 and Stage~2 policy networks are multilayer perceptrons with 3 hidden layers of 512, 256, and 128 units using ELU activations (Supplementary Table~1). PPO is used unmodified in both stages.

During training, we reset the environment if the base or a thigh contacts the terrain, or if the episode length reaches 20\,s. We employ a terrain curriculum from easy to difficult: slopes (0--0.4\,rad), stairs (0.05--0.25\,m height, 0.3\,m width), gaps (0.05--0.55\,m width, with distance between gaps from 0.7 to 0.25\,m), rough uneven (flat to $\pm$0.1\,m linear tessellation), and random discrete boxes (0.05--0.5\,m height offsets from randomly placed 1--2\,m cubes). New velocity commands between 0.4 and 2 m/s are sampled every 5\,s. During Stage~1, gaits are randomly re-sampled every few seconds, which also serves to expose the policy to forced transitions. Every 15\,s, a random horizontal base-velocity perturbation of up to 0.5 m/s is applied to the base. The policy outputs modulation signals at 33\,Hz, and the joint position targets from inverse kinematics are tracked by a PD controller at 1\,kHz. Equations~\ref{eq:r} and~\ref{eq:theta} are also integrated at 1\,kHz.

The heightmap encoder is trained with DAgger\cite{ross2011dagger} with the Stage~1 and Stage~2 networks frozen. At each iteration the encoder is optimized to regress the ground-truth height-map from the depth and proprioceptive stream under a mean-squared-error loss, while training states are collected on-policy from rollouts in which the controller is driven by the encoder's own predictions, so that the training distribution matches deployment and prediction errors do not compound. Distillation spans the full terrain curriculum. Depth images are rendered with the camera intrinsics, mounting pose and field of view matched to the hardware sensor, and corrupted with sensor-realistic perturbations (multiplicative range noise, random pixel dropout, and a per-episode calibration bias) for robustness to the statistics of the real depth stream.

\subsection*{Evaluation metrics}

We evaluate performance with respect to the following metrics, computed over 100 robots traversing 10\,m terrains. Except for success rate, all metrics are computed only for robots that successfully complete the terrain.

\begin{itemize}
    \item \textbf{Success Rate}: the percent of robots that successfully reach the end of the terrain without falling.
    \item \textbf{Cost of Transport (COT)}: the COT is computed with $\frac{|P|}{m \, g \, v}$, where $P$ is the average power, $m$ is the mass of the system, $g$ is gravitational acceleration, and $v$ is the average velocity. While this metric is usually used for locomotion on flat terrain, this dimensionless value is still useful on varying terrain to provide relative comparison of the energy-efficiency of different gaits. 
    \item \textbf{Mean Angular Velocity}: the mean angular velocity of the base is computed with $\frac{1}{3} (|\omega_x| + |\omega_y| + |\omega_z| )$, which gives an approximate stability measure of the base. 
    \item \textbf{Mean Joint Acceleration}: the mean joint acceleration is computed as $\frac{1}{12}\sum_{i=1}^{12}|\ddot{q}_i|$, which gives an approximate measure of how much a particular gait necessitates fast motor changes to locomote in a given environment.
    \item \textbf{Mean CPG Amplitude}: the mean oscillator amplitude is computed as $\frac{1}{4}\sum_{i=1}^{4}|r_i|$, which gives an approximate correspondence to the average step length.
    \item \textbf{Mean CPG Frequency}: the mean oscillator frequency in Hz is computed as $\frac{1}{4}\sum_{i=1}^{4}\dot{\theta}_i/(2\pi)$, which gives an approximate correspondence to the average step frequency.
    \item \textbf{Mean Body Height}: the mean body height $h$ measures how high the policy sets the nominal body position on a given terrain.
    \item \textbf{Mean Ground Clearance}: the mean ground clearance $g_c$ measures how high the policy sets the swing foot ground clearance on a given terrain. 
    \item \textbf{Mean Foot $x$ offset}: the mean foot x-offset $x_{off}$ measures the fore-aft offset of the foot trajectory relative to the hip during locomotion on a given terrain. 
    \item \textbf{Gait-tracking Accuracy}: the gait-tracking accuracy is defined as the fraction of commanded CPG stance time during which the corresponding foot is in contact with the terrain, measuring how reliably the prescribed stance phases are realized as ground contacts (Supplementary Figure 2).
\end{itemize}

\section*{Acknowledgements}
The authors would like to thank François Longchamp for support with the camera mounts. The authors acknowledge using generative AI tools (Claude, ChatGPT) to assist with grammar editing and rephrasing certain parts of the text for clarity. 

\section*{Funding Statement}
This research is partially supported by the Swiss National Science Foundation (SNSF) as part of project No.197237. 

\section*{Author Contributions}
G.B.: conceptualization, formal analysis, methodology, software, simulation, investigation, writing original draft, review \& editing, visualization;  A.I.: conceptualization, formal analysis, review \& editing, funding acquisition.

\section*{Competing Interests}
The authors declare no competing interests.

\bibliographystyle{naturemag}
\bibliography{refs}

@article{hildebrand1965symmetrical,
  title={Symmetrical gaits of horses},
  author={Hildebrand, Milton},
  journal={Science},
  volume={150},
  number={3697},
  pages={701--708},
  year={1965},
  publisher={American Association for the Advancement of Science},
  doi = {10.1126/science.150.3697.701},
  URL = {https://www.science.org/doi/abs/10.1126/science.150.3697.701}
}

@article{hildebrand1977analysis,
  title={Analysis of asymmetrical gaits},
  author={Hildebrand, Milton},
  journal={Journal of Mammalogy},
  volume={58},
  number={2},
  pages={131--156},
  year={1977},
  publisher={Oxford University Press},
  doi = {10.2307/1379571},
  url = {https://doi.org/10.2307/1379571}
}

@article{alexander1984gaits,
  title={The gaits of bipedal and quadrupedal animals},
  author={Alexander, R McN},
  journal={The International Journal of Robotics Research},
  volume={3},
  number={2},
  pages={49--59},
  year={1984},
  publisher={Sage Publications},
  doi = {10.1177/027836498400300205},
  URL = {https://doi.org/10.1177/027836498400300205}
}

@article{grillner2020current,
  title={Current principles of motor control, with special reference to vertebrate locomotion},
  author={Grillner, Sten and El Manira, Abdeljabbar},
  journal={Physiological Reviews},
  volume={100},
  number={1},
  pages={271--320},
  year={2020},
  publisher={American Physiological Society Bethesda, MD},
  doi = {10.1152/physrev.00015.2019},
  URL = {https://doi.org/10.1152/physrev.00015.2019}
}

@article{hoyt1981gait,
  title={Gait and the energetics of locomotion in horses},
  author={Hoyt, Donald F and Taylor, C Richard},
  journal={Nature},
  volume={292},
  number={5820},
  pages={239--240},
  year={1981},
  publisher={Nature Publishing Group},
  doi = {10.1038/292239a0},
  url={https://doi.org/10.1038/292239a0}
}

@article{shafiee2024viability,
  title={Viability leads to the emergence of gait transitions in learning agile quadrupedal locomotion on challenging terrains},
  author={Shafiee, Milad and Bellegarda, Guillaume and Ijspeert, Auke},
  journal={Nature Communications},
  volume={15},
  number={1},
  pages={3073},
  year={2024},
  publisher={Nature Publishing Group UK London},
  doi = {10.1038/s41467-024-47443-w},
  url={https://doi.org/10.1038/s41467-024-47443-w}
}

@inproceedings{cheng2024extreme,
  title={Extreme parkour with legged robots},
  author={Cheng, Xuxin and Shi, Kexin and Agarwal, Ananye and Pathak, Deepak},
  booktitle={2024 IEEE International Conference on Robotics and Automation (ICRA)},
  pages={11443--11450},
  year={2024},
  organization={IEEE},
  doi = {10.1109/ICRA57147.2024.10610200},
}

@InProceedings{zhuang2023robot,
  title = 	 {Robot Parkour Learning},
  author =       {Zhuang, Ziwen and Fu, Zipeng and Wang, Jianren and Atkeson, Christopher G. and Schwertfeger, S\"{o}ren and Finn, Chelsea and Zhao, Hang},
  booktitle = 	 {Proceedings of The 7th Conference on Robot Learning},
  pages = 	 {73--92},
  year = 	 {2023},
  editor = 	 {Tan, Jie and Toussaint, Marc and Darvish, Kourosh},
  volume = 	 {229},
  series = 	 {Proceedings of Machine Learning Research},
  month = 	 {06--09 Nov},
  publisher =    {PMLR},
  url = 	 {https://proceedings.mlr.press/v229/zhuang23a.html},
}

@article{hoeller2024anymal,
  title={{ANY}mal parkour: Learning agile navigation for quadrupedal robots},
  author={Hoeller, David and Rudin, Nikita and Sako, Dhionis and Hutter, Marco},
  journal={Science Robotics},
  volume={9},
  number={88},
  pages={eadi7566},
  year={2024},
  publisher={American Association for the Advancement of Science},
  doi = {10.1126/scirobotics.adi7566},
  URL = {https://www.science.org/doi/abs/10.1126/scirobotics.adi7566}
}

@article{lee2020learning,
  title={Learning quadrupedal locomotion over challenging terrain},
  author={Lee, Joonho and Hwangbo, Jemin and Wellhausen, Lorenz and Koltun, Vladlen and Hutter, Marco},
  journal={Science Robotics},
  volume={5},
  number={47},
  pages={eabc5986},
  year={2020},
  publisher={American Association for the Advancement of Science},
  doi = {10.1126/scirobotics.abc5986},
  URL = {https://www.science.org/doi/abs/10.1126/scirobotics.abc5986}
}

@article{
miki2022learning,
author = {Takahiro Miki  and Joonho Lee  and Jemin Hwangbo  and Lorenz Wellhausen  and Vladlen Koltun  and Marco Hutter },
title = {Learning robust perceptive locomotion for quadrupedal robots in the wild},
journal = {Science Robotics},
volume = {7},
number = {62},
pages = {eabk2822},
year = {2022},
doi = {10.1126/scirobotics.abk2822},
URL = {https://www.science.org/doi/abs/10.1126/scirobotics.abk2822}
}

@INPROCEEDINGS{bellegarda2022robust,
  author={Bellegarda, Guillaume and Chen, Yiyu and Liu, Zhuochen and Nguyen, Quan},
  booktitle={2022 IEEE/RSJ International Conference on Intelligent Robots and Systems (IROS)}, 
  title={Robust High-Speed Running for Quadruped Robots via Deep Reinforcement Learning}, 
  year={2022},
  volume={},
  number={},
  pages={10364-10370},
  doi={10.1109/IROS47612.2022.9982132}}

@article{margolis2022rapid,
author = {Gabriel B. Margolis and Ge Yang and Kartik Paigwar and Tao Chen and Pulkit Agrawal},
title ={Rapid locomotion via reinforcement learning},
journal = {The International Journal of Robotics Research},
volume = {43},
number = {4},
pages = {572-587},
year = {2024},
doi = {10.1177/02783649231224053},
URL = {https://doi.org/10.1177/02783649231224053},
}

@article{ji2022concurrent,
  title={Concurrent Training of a Control Policy and a State Estimator for Dynamic and Robust Legged Locomotion},
  author={Ji, Gwanghyeon and Mun, Juhyeok and Kim, Hyeongjun and Hwangbo, Jemin},
  journal={IEEE Robotics and Automation Letters},
  volume={7},
  number={2},
  pages={4630--4637},
  year={2022},
  publisher={IEEE},
  doi = {10.1109/LRA.2022.3151396},
}

@InProceedings{fu2022energy,
  title = 	 {Minimizing Energy Consumption Leads to the Emergence of Gaits in Legged Robots},
  author =       {Fu, Zipeng and Kumar, Ashish and Malik, Jitendra and Pathak, Deepak},
  booktitle = 	 {Proceedings of the 5th Conference on Robot Learning},
  pages = 	 {928--937},
  year = 	 {2022},
  volume = 	 {164},
  series = 	 {Proceedings of Machine Learning Research},
  month = 	 {08--11 Nov},
  publisher =    {PMLR},
  url = 	 {https://proceedings.mlr.press/v164/fu22a.html},
}

@InProceedings{rudin2022anymalisaac,
  title = 	 {Learning to Walk in Minutes Using Massively Parallel Deep Reinforcement Learning},
  author =       {Rudin, Nikita and Hoeller, David and Reist, Philipp and Hutter, Marco},
  booktitle = 	 {Proceedings of the 5th Conference on Robot Learning},
  pages = 	 {91--100},
  year = 	 {2022},
  volume = 	 {164},
  series = 	 {Proceedings of Machine Learning Research},
  month = 	 {08--11 Nov},
  publisher =    {PMLR},
  url = 	 {https://proceedings.mlr.press/v164/rudin22a.html},
}

@Article{yu2023identifying,
author={Yu, Wanming
and Yang, Chuanyu
and McGreavy, Christopher
and Triantafyllidis, Eleftherios
and Bellegarda, Guillaume
and Shafiee, Milad
and Ijspeert, Auke Jan
and Li, Zhibin},
title={Identifying important sensory feedback for learning locomotion skills},
journal={Nature Machine Intelligence},
year={2023},
month={Aug},
day={01},
volume={5},
number={8},
pages={919-932},
issn={2522-5839},
doi={10.1038/s42256-023-00701-w},
url={https://doi.org/10.1038/s42256-023-00701-w}
}

@inproceedings{yang2024agile,
  title={Agile continuous jumping in discontinuous terrains},
  author={Yang, Yuxiang and Shi, Guanya and Lin, Changyi and Meng, Xiangyun and Scalise, Rosario and Castro, Mateo Guaman and Yu, Wenhao and Zhang, Tingnan and Zhao, Ding and Tan, Jie and others},
  booktitle={2025 IEEE International Conference on Robotics and Automation (ICRA)},
  pages={10245--10252},
  year={2025},
  organization={IEEE},
  doi = {10.1109/ICRA55743.2025.11127314},
}

@INPROCEEDINGS{peng2020learning, 
    AUTHOR    = {Xue Bin Peng AND Erwin Coumans AND Tingnan Zhang AND Tsang-Wei Lee AND Jie Tan AND Sergey Levine}, 
    TITLE     = {{Learning Agile Robotic Locomotion Skills by Imitating Animals}}, 
    BOOKTITLE = {Proceedings of Robotics: Science and Systems}, 
    YEAR      = {2020}, 
    ADDRESS   = {Corvalis, Oregon, USA}, 
    MONTH     = {July}, 
    DOI       = {10.15607/RSS.2020.XVI.064} 
}

@inproceedings{escontrela2022adversarial,
  title={Adversarial motion priors make good substitutes for complex reward functions},
  author={Escontrela, Alejandro and Peng, Xue Bin and Yu, Wenhao and Zhang, Tingnan and Iscen, Atil and Goldberg, Ken and Abbeel, Pieter},
  booktitle={2022 IEEE/RSJ International Conference on Intelligent Robots and Systems (IROS)},
  pages={25--32},
  year={2022},
  organization={IEEE},
  doi = {10.1109/IROS47612.2022.9981973},
}

@article{han2024lifelike,
  title={Lifelike agility and play in quadrupedal robots using reinforcement learning and generative pre-trained models},
  author={Han, Lei
and Zhu, Qingxu
and Sheng, Jiapeng
and Zhang, Chong
and Li, Tingguang
and Zhang, Yizheng
and Zhang, He
and Liu, Yuzhen
and Zhou, Cheng
and Zhao, Rui
and Li, Jie
and Zhang, Yufeng
and Wang, Rui
and Chi, Wanchao
and Li, Xiong
and Zhu, Yonghui
and Xiang, Lingzhu
and Teng, Xiao
and Zhang, Zhengyou},
  journal={Nature Machine Intelligence},
  volume={6},
  number={7},
  pages={787--798},
  year={2024},
  publisher={Nature Publishing Group UK London},
  doi = {10.1038/s42256-024-00861-3},
  url={https://doi.org/10.1038/s42256-024-00861-3}
}

@Article{fu2026learning,
author={Fu, Huiqiao
and Dong, Haoyu
and Xu, Wentao
and Zhou, Zhehao
and Deng, Guizhou
and Tang, Kaiqiang
and Dong, Daoyi
and Chen, Chunlin},
title={Learning diverse natural behaviors for enhancing the agility of quadrupedal robots},
journal={Nature Communications},
year={2026},
month={Apr},
day={29},
volume={17},
number={1},
pages={5857},
doi={10.1038/s41467-026-72475-9},
url={https://doi.org/10.1038/s41467-026-72475-9}
}

@INPROCEEDINGS{dicarlo2018mpc,
  author={J. {Di Carlo} and P. M. {Wensing} and B. {Katz} and G. {Bledt} and S. {Kim}},
  booktitle={2018 IEEE/RSJ International Conference on Intelligent Robots and Systems (IROS)}, 
  title={Dynamic Locomotion in the {MIT} Cheetah 3 Through Convex Model-Predictive Control}, 
  year={2018},
  volume={},
  number={},
  pages={1-9},
  doi = {10.1109/IROS.2018.8594448},
}

@article{kim2019highly,
  title={Highly dynamic quadruped locomotion via whole-body impulse control and model predictive control},
  author={Kim, Donghyun and Di Carlo, Jared and Katz, Benjamin and Bledt, Gerardo and Kim, Sangbae},
  journal={arXiv preprint arXiv:1909.06586},
  year={2019},
  url = {https://arxiv.org/abs/1909.06586},
}

@ARTICLE{bellicoso2018dynamic,
  author={C. D. {Bellicoso} and F. {Jenelten} and C. {Gehring} and M. {Hutter}},
  journal={IEEE Robotics and Automation Letters}, 
  title={Dynamic Locomotion Through Online Nonlinear Motion Optimization for Quadrupedal Robots}, 
  year={2018},
  volume={3},
  number={3},
  pages={2261-2268},
  doi = {10.1109/LRA.2018.2794620},
}

@ARTICLE{winkler2018gait,
  author={Winkler, Alexander W. and Bellicoso, C. Dario and Hutter, Marco and Buchli, Jonas},
  journal={IEEE Robotics and Automation Letters}, 
  title={Gait and Trajectory Optimization for Legged Systems Through Phase-Based End-Effector Parameterization}, 
  year={2018},
  volume={3},
  number={3},
  pages={1560-1567},
  doi={10.1109/LRA.2018.2798285}
  }

@ARTICLE{aceituno2018simultaneous,
  author={Aceituno-Cabezas, Bernardo and Mastalli, Carlos and Dai, Hongkai and Focchi, Michele and Radulescu, Andreea and Caldwell, Darwin G. and Cappelletto, José and Grieco, Juan C. and Fernández-López, Gerardo and Semini, Claudio},
  journal={IEEE Robotics and Automation Letters}, 
  title={Simultaneous Contact, Gait, and Motion Planning for Robust Multilegged Locomotion via Mixed-Integer Convex Optimization}, 
  year={2018},
  volume={3},
  number={3},
  pages={2531-2538},
  doi={10.1109/LRA.2017.2779821}
  }

@article{jenelten2022tamols,
  title={{TAMOLS}: Terrain-aware motion optimization for legged systems},
  author={Jenelten, Fabian and Grandia, Ruben and Farshidian, Farbod and Hutter, Marco},
  journal={IEEE Transactions on Robotics},
  year={2022},
  publisher={IEEE},
  volume = {38},
  number = {6},
  pages = {3395--3413},
  doi = {10.1109/TRO.2022.3186804},
}

@article{jenelten2024dtc,
  title={{DTC}: Deep Tracking Control},
  author={Jenelten, Fabian and He, Junzhe and Farshidian, Farbod and Hutter, Marco},
  journal={Science Robotics},
  volume={9},
  number={86},
  pages={eadh5401},
  year={2024},
  publisher={American Association for the Advancement of Science},
  doi = {10.1126/scirobotics.adh5401},
  URL = {https://www.science.org/doi/abs/10.1126/scirobotics.adh5401},
}

@InProceedings{yang2022fast,
  title = 	 {Fast and Efficient Locomotion via Learned Gait Transitions},
  author =       {Yang, Yuxiang and Zhang, Tingnan and Coumans, Erwin and Tan, Jie and Boots, Byron},
  booktitle = 	 {Proceedings of the 5th Conference on Robot Learning},
  pages = 	 {773--783},
  year = 	 {2022},
  volume = 	 {164},
  series = 	 {Proceedings of Machine Learning Research},
  month = 	 {08--11 Nov},
  publisher =    {PMLR},
  url = 	 {https://proceedings.mlr.press/v164/yang22d.html},
}

@INPROCEEDINGS{wang2020automatic,
  author={Wang, Jiayi and Chatzinikolaidis, Iordanis and Mastalli, Carlos and Wolfslag, Wouter and Xin, Guiyang and Tonneau, Steve and Vijayakumar, Sethu},
  booktitle={2020 IEEE/RSJ International Conference on Intelligent Robots and Systems (IROS)}, 
  title={Automatic Gait Pattern Selection for Legged Robots}, 
  year={2020},
  volume={},
  number={},
  pages={3990-3997},
  doi={10.1109/IROS45743.2020.9340789}}

@ARTICLE{luo2025tfgait,
  author={Luo, Aocheng and Wan, Qifeng and Meng, Yan and Kong, Shihan and Chi, Wanchao and Zhang, Chong and Zhang, Shenghao and Liu, Yuzhen and Zhu, Qiuguo and Yu, Junzhi},
  journal={IEEE Transactions on Automation Science and Engineering}, 
  title={{TFGait}—Stable and Efficient Adaptive Gait Planning With Terrain Recognition and {Froude} Number for Quadruped Robot}, 
  year={2025},
  volume={22},
  number={},
  pages={9218-9229},
  doi={10.1109/TASE.2024.3503277}}

@InProceedings{margolis2022pixels,
  title = 	 {Learning to Jump from Pixels},
  author =       {Margolis, Gabriel B and Chen, Tao and Paigwar, Kartik and Fu, Xiang and Kim, Donghyun and Kim, Sang bae and Agrawal, Pulkit},
  booktitle = 	 {Proceedings of the 5th Conference on Robot Learning},
  pages = 	 {1025--1034},
  year = 	 {2022},
  volume = 	 {164},
  series = 	 {Proceedings of Machine Learning Research},
  month = 	 {08--11 Nov},
  publisher =    {PMLR},
  url = 	 {https://proceedings.mlr.press/v164/margolis22a.html},
}

@InProceedings{yang2023cajun,
  title = 	 {{CAJ}un: Continuous Adaptive Jumping using a Learned Centroidal Controller},
  author =       {Yang, Yuxiang and Shi, Guanya and Meng, Xiangyun and Yu, Wenhao and Zhang, Tingnan and Tan, Jie and Boots, Byron},
  booktitle = 	 {Proceedings of The 7th Conference on Robot Learning},
  pages = 	 {2791--2806},
  year = 	 {2023},
  volume = 	 {229},
  series = 	 {Proceedings of Machine Learning Research},
  month = 	 {06--09 Nov},
  publisher =    {PMLR},
  url = 	 {https://proceedings.mlr.press/v229/yang23b.html},
}

@article{ijspeert2008,
title = {Central pattern generators for locomotion control in animals and robots: A review},
journal = {Neural Networks},
volume = {21},
number = {4},
pages = {642-653},
year = {2008},
issn = {0893-6080},
doi = {10.1016/j.neunet.2008.03.014},
author = {Auke Jan Ijspeert},
url = {https://www.sciencedirect.com/science/article/pii/S0893608008000804},
}

@article{bellegarda2022cpgrl,
  title={{CPG-RL}: Learning Central Pattern Generators for Quadruped Locomotion},
  author={Bellegarda, Guillaume and Ijspeert, Auke},
  journal={IEEE Robotics and Automation Letters},
  volume={7},
  number={4},
  pages={12547--12554},
  year={2022},
  publisher={IEEE},
  doi = {10.1109/LRA.2022.3218167},
}

@article{von1950reafferenzprinzip,
  title={Das reafferenzprinzip: Wechselwirkungen zwischen zentralnervensystem und peripherie},
  author={Von Holst, Erich and Mittelstaedt, Horst},
  journal={Naturwissenschaften},
  volume={37},
  number={20},
  pages={464--476},
  year={1950},
  publisher={Springer},
  doi = {10.1007/BF00622503},
}

@article{crapse2008corollary,
  title={Corollary discharge across the animal kingdom},
  author={Crapse, Trinity B and Sommer, Marc A},
  journal={Nature Reviews Neuroscience},
  volume={9},
  number={8},
  pages={587--600},
  year={2008},
  publisher={Nature Publishing Group UK London},
  doi = {10.1038/nrn2457},
}

@inproceedings{bellegarda2024visualcpgrl,
  title={Visual {CPG-RL}: Learning Central Pattern Generators for Visually-Guided Quadruped Locomotion},
  author={Bellegarda, Guillaume and Shafiee, Milad and Ijspeert, Auke},
  pages={1420-1427},
  year={2024},
  publisher={IEEE},
  doi = {10.1109/ICRA57147.2024.10611128},
  booktitle = {2024 IEEE International Conference on Robotics and Automation (ICRA)},
}

@article{owaki2017quadruped,
  title={A quadruped robot exhibiting spontaneous gait transitions from walking to trotting to galloping},
  author={Owaki, Dai and Ishiguro, Akio},
  journal={Scientific reports},
  volume={7},
  number={1},
  pages={277},
  year={2017},
  publisher={Nature Publishing Group},
  doi = {10.1038/s41598-017-00348-9},
}

@INPROCEEDINGS{zhang2024learning,
  author={Zhang, Jenny and Heim, Steve and Jeon, Se Hwan and Kim, Sangbae},
  booktitle={2024 IEEE International Conference on Robotics and Automation (ICRA)}, 
  title={Learning Emergent Gaits with Decentralized Phase Oscillators: on the role of Observations, Rewards, and Feedback}, 
  year={2024},
  volume={},
  number={},
  pages={3426-3433},
  doi={10.1109/ICRA57147.2024.10611045}}

@INPROCEEDINGS{zhang2024online,
  author={Zhang, Zewei and Bellegarda, Guillaume and Shafiee, Milad and Ijspeert, Auke},
  booktitle={2024 IEEE/RSJ International Conference on Intelligent Robots and Systems (IROS)}, 
  title={Online Optimization of Central Pattern Generators for Quadruped Locomotion}, 
  year={2024},
  volume={},
  number={},
  pages={13547-13554},
  doi={10.1109/IROS58592.2024.10802762}}

@INPROCEEDINGS{bellegarda2024quadruped,
  author={Bellegarda, Guillaume and Shafiee, Milad and Özberk, Merih Ekin and Ijspeert, Auke},
  booktitle={2024 IEEE International Conference on Robotics and Automation (ICRA)}, 
  title={Quadruped-{F}rog: Rapid Online Optimization of Continuous Quadruped Jumping}, 
  year={2024},
  volume={},
  number={},
  pages={1443-1450},
  doi={10.1109/ICRA57147.2024.10610141}}

@inproceedings{bellegarda2025allgaits,
  title={All{G}aits: Learning all quadruped gaits and transitions},
  author={Bellegarda, Guillaume and Shafiee, Milad and Ijspeert, Auke},
  booktitle={2025 IEEE International Conference on Robotics and Automation (ICRA)},
  pages={15929--15935},
  year={2025},
  organization={IEEE},
  doi = {10.1109/ICRA55743.2025.11127285},
}

@InProceedings{margolis2022walktheseways,
  title = 	 {Walk These Ways: Tuning Robot Control for Generalization with Multiplicity of Behavior},
  author =       {Margolis, Gabriel B. and Agrawal, Pulkit},
  booktitle = 	 {Proceedings of The 6th Conference on Robot Learning},
  pages = 	 {22--31},
  year = 	 {2023},
  volume = 	 {205},
  series = 	 {Proceedings of Machine Learning Research},
  month = 	 {14--18 Dec},
  publisher =    {PMLR},
  url = 	 {https://proceedings.mlr.press/v205/margolis23a.html},
}

@ARTICLE{shao2022gait,
  author={Shao, Yecheng and Jin, Yongbin and Liu, Xianwei and He, Weiyan and Wang, Hongtao and Yang, Wei},
  journal={IEEE Robotics and Automation Letters}, 
  title={Learning Free Gait Transition for Quadruped Robots Via Phase-Guided Controller}, 
  year={2022},
  volume={7},
  number={2},
  pages={1230-1237},
  doi={10.1109/LRA.2021.3136645}}

@ARTICLE{liu2025skill,
  author={Liu, Xin and Wu, Jinze and Xue, Yufei and Qi, Chenkun and Xin, Guiyang and Gao, Feng},
  journal={IEEE Transactions on Industrial Electronics}, 
  title={Skill Latent Space Based Multigait Learning for a Legged Robot}, 
  year={2025},
  volume={72},
  number={2},
  pages={1743-1752},
  doi={10.1109/TIE.2024.3429576}}

@InProceedings{mitchell2025gaitor,
  title = 	 {Gaitor: Learning a Unified Representation Across Gaits for Real-World Quadruped Locomotion},
  author =       {Mitchell, Alexander Luis and Merkt, Wolfgang and Papatheodorou, Aristotelis and Havoutis, Ioannis and Posner, Ingmar},
  booktitle = 	 {Proceedings of The 8th Conference on Robot Learning},
  pages = 	 {780--793},
  year = 	 {2025},
  volume = 	 {270},
  series = 	 {Proceedings of Machine Learning Research},
  month = 	 {06--09 Nov},
  publisher =    {PMLR},
  url = 	 {https://proceedings.mlr.press/v270/mitchell25a.html},
}

@ARTICLE{stamatopoulou2025egait,
  author={Stamatopoulou, Maria and Tan, Daniel and Bendikas, Rokas and Modugno, Valerio and Li, Zhibin and Kanoulas, Dimitrios},
  journal={IEEE Transactions on Automation Science and Engineering}, 
  title={{eGAIT}: Multi-Skilled Policy for Energy-Efficient Gait Transitions}, 
  year={2026},
  volume={23},
  number={},
  pages={1575-1586},
  doi={10.1109/TASE.2025.3636224}}

@Article{humphreys2025learning,
    title   = {Learning to adapt through bio-inspired gait strategies for versatile quadruped locomotion},
    author  = {Humphreys, Joseph and Zhou, Chengxu},
    journal = {Nature Machine Intelligence},
    volume  = {7},
    pages   = {1141--1153},
    year    = {2025},
    issn    = {2522-5839},
    doi     = {10.1038/s42256-025-01065-z},
    url     = {https://doi.org/10.1038/s42256-025-01065-z}
}

@article{
kang2026agile,
author = {Jun-Gill Kang  and Jaehyun Park  and Tae-Gyu Song  and Joon-Ha Kim  and Seungwoo Hong  and Hae-Won Park },
title = {Agile perceptive multiskill locomotion for quadrupedal robots in the wild},
journal = {Science Robotics},
volume = {11},
number = {116},
pages = {eadz7397},
year = {2026},
doi = {10.1126/scirobotics.adz7397},
URL = {https://www.science.org/doi/abs/10.1126/scirobotics.adz7397}
}

@inproceedings{shafiee2024manyquadrupeds,
  title={Many{Q}uadrupeds: Learning a single locomotion policy for diverse quadruped robots},
  author={Shafiee, Milad and Bellegarda, Guillaume and Ijspeert, Auke},
  booktitle={2024 IEEE International Conference on Robotics and Automation (ICRA)},
  pages={3471--3477},
  year={2024},
  organization={IEEE},
  doi = {10.1109/ICRA57147.2024.10610155},
}

@inproceedings{kim2025stage,
  title={Stage-wise reward shaping for acrobatic robots: A constrained multi-objective reinforcement learning approach},
  author={Kim, Dohyeong and Kwon, Hyeokjin and Kim, Junseok and Lee, Gunmin and Oh, Songhwai},
  booktitle={2025 IEEE International Conference on Robotics and Automation (ICRA)},
  pages={10268--10274},
  year={2025},
  organization={IEEE},
  doi = {10.1109/ICRA55743.2025.11128552},
}

@article{xiao2025learning,
  title={Learning stable bipedal locomotion skills for quadrupedal robots on challenging terrains with automatic fall recovery},
  author={Xiao, Erdong and Dong, Yinzhao and Lam, James and Lu, Peng},
  journal={npj Robotics},
  volume={3},
  number={1},
  pages={22},
  year={2025},
  publisher={Nature Publishing Group UK London},
  doi = {10.1038/s44182-025-00043-2},
  url={https://doi.org/10.1038/s44182-025-00043-2}
}

@INPROCEEDINGS{zakka2025mujocoplayground, 
    AUTHOR    = {Kevin Zakka AND Baruch Tabanpour AND Qiayuan Liao AND Mustafa Haiderbhai AND Samuel Holt AND Jing Yuan Luo AND Arthur Allshire AND Erik Frey AND Koushil Sreenath AND Lueder Alexander Kahrs AND Carmelo Sferrazza AND Yuval Tassa AND Pieter Abbeel}, 
    TITLE     = {{Demonstrating MuJoCo Playground}}, 
    BOOKTITLE = {Proceedings of Robotics: Science and Systems}, 
    YEAR      = {2025}, 
    ADDRESS   = {Los Angeles, CA, USA}, 
    MONTH     = {June}, 
    DOI       = {10.15607/RSS.2025.XXI.020} 
}

@INPROCEEDINGS{fuchioka2023optmimic,
  author={Fuchioka, Yuni and Xie, Zhaoming and Van de Panne, Michiel},
  booktitle={2023 IEEE International Conference on Robotics and Automation (ICRA)}, 
  title={{OPT-Mimic}: Imitation of Optimized Trajectories for Dynamic Quadruped Behaviors}, 
  year={2023},
  volume={},
  number={},
  pages={5092-5098},
  doi={10.1109/ICRA48891.2023.10160562}}

@INPROCEEDINGS{smith2023learning, 
    AUTHOR    = {Laura M Smith AND J. Chase Kew AND Tianyu Li AND Linda Luu AND Xue Bin Peng AND Sehoon Ha AND Jie Tan AND Sergey Levine}, 
    TITLE     = {{Learning and Adapting Agile Locomotion Skills by Transferring Experience}}, 
    BOOKTITLE = {Proceedings of Robotics: Science and Systems}, 
    YEAR      = {2023}, 
    ADDRESS   = {Daegu, Republic of Korea}, 
    MONTH     = {July}, 
    DOI       = {10.15607/RSS.2023.XIX.051} 
}

@INPROCEEDINGS{tu2025continuous,
  author={Tu, Jiaxin and Wei, Xiaoyi and Zhang, Yueqi and Hou, Taixian and Gao, Xiaofei and Dong, Zhiyan and Zhai, Peng and Zhang, Lihua},
  booktitle={2025 IEEE International Conference on Robotics and Automation (ICRA)}, 
  title={Continuous Control of Diverse Skills in Quadruped Robots Without Complete Expert Datasets}, 
  year={2025},
  volume={},
  number={},
  pages={11191-11197},
  doi={10.1109/ICRA55743.2025.11128738}}

@article{farley1991mechanical,
  title={A mechanical trigger for the trot-gallop transition in horses},
  author={Farley, Claire T and Taylor, C Richard},
  journal={Science},
  volume={253},
  number={5017},
  pages={306--308},
  year={1991},
  publisher={American Association for the Advancement of Science},
  doi = {10.1126/science.1857965},
  URL = {https://www.science.org/doi/abs/10.1126/science.1857965},
}

@article{carlsonkuhta1998upslope,
  title={Forms of forward quadrupedal locomotion. II. A comparison of posture, hindlimb kinematics, and motor patterns for upslope and level walking},
  author={Carlson-Kuhta, Patricia and Trank, Tamara V and Smith, Judith L},
  journal={Journal of Neurophysiology},
  volume={79},
  number={4},
  pages={1687--1701},
  year={1998},
  publisher={American Physiological Society Bethesda, MD},
  doi = {10.1152/jn.1998.79.4.1687},
  URL = {https://doi.org/10.1152/jn.1998.79.4.1687},
}

@article{smith1998downslope,
  title={Forms of forward quadrupedal locomotion. III. A comparison of posture, hindlimb kinematics, and motor patterns for downslope and level walking},
  author={Smith, Judith L and Carlson-Kuhta, Patricia and Trank, Tamara V},
  journal={Journal of Neurophysiology},
  volume={79},
  number={4},
  pages={1702--1716},
  year={1998},
  publisher={American Physiological Society Bethesda, MD},
  doi = {10.1152/jn.1998.79.4.1702},
  URL = {https://doi.org/10.1152/jn.1998.79.4.1702},
}

@article{vanbijlert2024muscle,
  title={Muscle-driven predictive physics simulations of quadrupedal locomotion in the horse},
  author={van Bijlert, Pasha A and Geijtenbeek, Thomas and Smit, Ineke H and Schulp, Anne S and Bates, Karl T},
  journal={Integrative and Comparative Biology},
  volume={64},
  number={3},
  pages={694--714},
  year={2024},
  publisher={Oxford University Press},
  doi = {10.1093/icb/icae095},
  url = {https://doi.org/10.1093/icb/icae095},
}

@article{berendes2016speed,
  title={Speed-dependent interplay between local pattern-generating activity and sensory signals during walking in {Drosophila}},
  author={Berendes, Volker and Zill, Sasha N and B{\"u}schges, Ansgar and Bockem{\"u}hl, Till},
  journal={Journal of Experimental Biology},
  volume={219},
  number={23},
  pages={3781--3793},
  year={2016},
  publisher={The Company of Biologists Ltd},
  doi = {10.1242/jeb.146720},
  url = {https://doi.org/10.1242/jeb.146720},
}

@article{makoviychuk2021isaac,
      title={Isaac Gym: High Performance GPU-Based Physics Simulation For Robot Learning}, 
      author={Viktor Makoviychuk and Lukasz Wawrzyniak and Yunrong Guo and Michelle Lu and Kier Storey and Miles Macklin and David Hoeller and Nikita Rudin and Arthur Allshire and Ankur Handa and Gavriel State},
      year={2021},
      journal={arXiv preprint arXiv:2108.10470},
      url = {https://arxiv.org/abs/2108.10470},
}

@article{schulman2017proximal,
  title={Proximal Policy Optimization Algorithms},
  author={Schulman, John and Wolski, Filip and Dhariwal, Prafulla and Radford, Alec and Klimov, Oleg},
  journal={arXiv preprint arXiv:1707.06347},
  year={2017},
  url       = {https://arxiv.org/abs/1707.06347},
}

@InProceedings{ross2011dagger,
  title = 	 {A Reduction of Imitation Learning and Structured Prediction to No-Regret Online Learning},
  author = 	 {Ross, Stephane and Gordon, Geoffrey and Bagnell, Drew},
  booktitle = 	 {Proceedings of the Fourteenth International Conference on Artificial Intelligence and Statistics},
  pages = 	 {627--635},
  year = 	 {2011},
  volume = 	 {15},
  series = 	 {Proceedings of Machine Learning Research},
  address = 	 {Fort Lauderdale, FL, USA},
  month = 	 {11--13 Apr},
  publisher =    {PMLR},
  url = 	 {https://proceedings.mlr.press/v15/ross11a.html},
}

\clearpage
\clearpage

\setcounter{figure}{0}
\setcounter{table}{0}

\renewcommand{\theHfigure}{supp.\arabic{figure}}
\renewcommand{\theHtable}{supp.\arabic{table}}

\captionsetup[figure]{
    name={Supplementary Fig.},
    font=footnotesize,
    labelsep=period,
    labelfont=bf
}

\captionsetup[table]{
    name={Supplementary Table},
    labelsep=period,
    labelfont=bf
}

\setcounter{secnumdepth}{0}
\setcounter{tocdepth}{2}

\begin{center}
    {\Large\bfseries Supplementary Information\par}
    \vspace{0.8em}
    {\large\bfseries
    Learning optimal gaits and styles for terrain-adaptive
    quadruped locomotion\par}
    \vspace{1em}
    {\normalsize Guillaume Bellegarda and Auke Ijspeert\par}
\end{center}

\tableofcontents
\newpage

\section{Gait definitions and tracking}
\subsection{Supplementary Figure 1: Gait definitions and coupling matrices}

Supplementary Fig.~1 shows contact timing for each foot as a percentage of a single gait cycle for the 11 gaits considered in this paper: Lateral Sequence Walk (L.S.~Walk), Diagonal Sequence Walk (D.S.~Walk), Amble, Trot, Pace, Pronk, Half-Bound, Bound, Canter, Transverse Gallop (T.~Gallop), Rotary Gallop (R.~Gallop). We also define a test-time-only gait where three legs are coupled together in phase, and one leg out of phase, which is not found in nature. The contact timing represents the percentage of a single gait cycle at which that foot comes into contact with the ground. These timings are converted to pairwise phase offsets in radians, which define the $4 \times 4$ coupling matrices $\bm{\Phi}$ used in Equation~2 (main text). The columns/rows denote the offsets between the respective limbs: Front Right (FR), Front Left (FL), Hind Right (HR), Hind Left (HL).

As two minimal examples, the pronk gait is a matrix of zeros, since all legs are synchronized together. For the trot gait, diagonal limbs are paired, and have a $\pi$ offset with the opposing diagonal pair, which can be written as follows:
\begin{equation}
    \bm{\Phi}_{\text{trot}} =
    \begin{bmatrix}
       0 & \pi & \pi & 0 \\
    -\pi &   0 &   0 & -\pi \\
    -\pi &   0 &   0 & -\pi \\
       0 & \pi & \pi & 0
    \end{bmatrix}
\end{equation}

\begin{figure}[H]
\centering
\includegraphics[width=\textwidth]{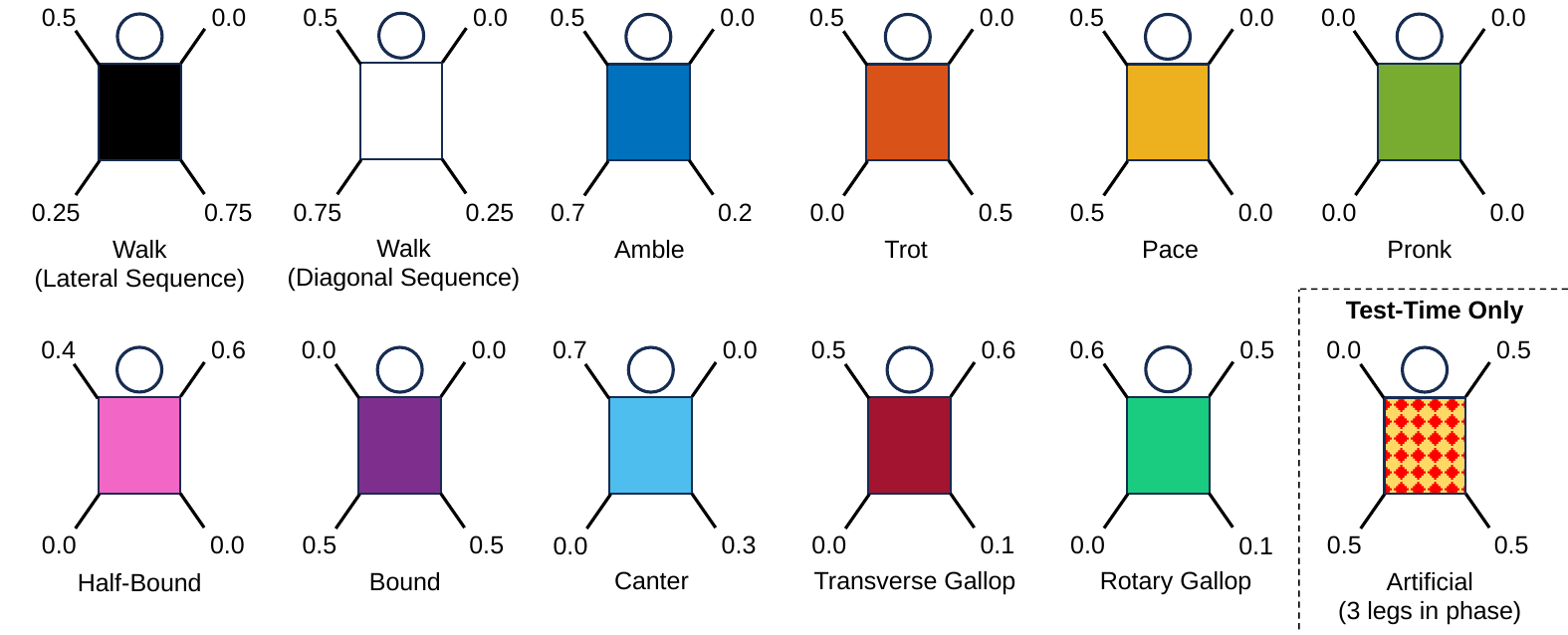}\\
\caption{Contact timing for each foot with the ground as a percentage of a single gait cycle for various quadruped gaits: Lateral Sequence Walk (L.S.~Walk), Diagonal Sequence Walk (D.S.~Walk), Amble, Trot, Pace, Pronk, Half-Bound, Bound, Canter, Transverse Gallop (T.~Gallop), Rotary Gallop (R.~Gallop). While not considered during training as it is not a biological gait, during test time we show that the policy can still adapt to an artificial gait where 3 legs are coupled together.
The timings are converted to matrices that denote phase offsets between different limbs (i.e.~0.5 corresponds to $\pi$) in column order: Front Right (FR), Front Left (FL), Hind Right (HR), Hind Left (HL), as they appear in Equation~2 (main text).
}
\label{fig:supp_gait_matrices}
\end{figure}

\newpage 

\subsection{Supplementary Figure 2: Gait-tracking accuracy across terrains and velocities}
\begin{figure}[H]
    \vspace{-0.5em}
    \centering
    \includegraphics[width=\textwidth]{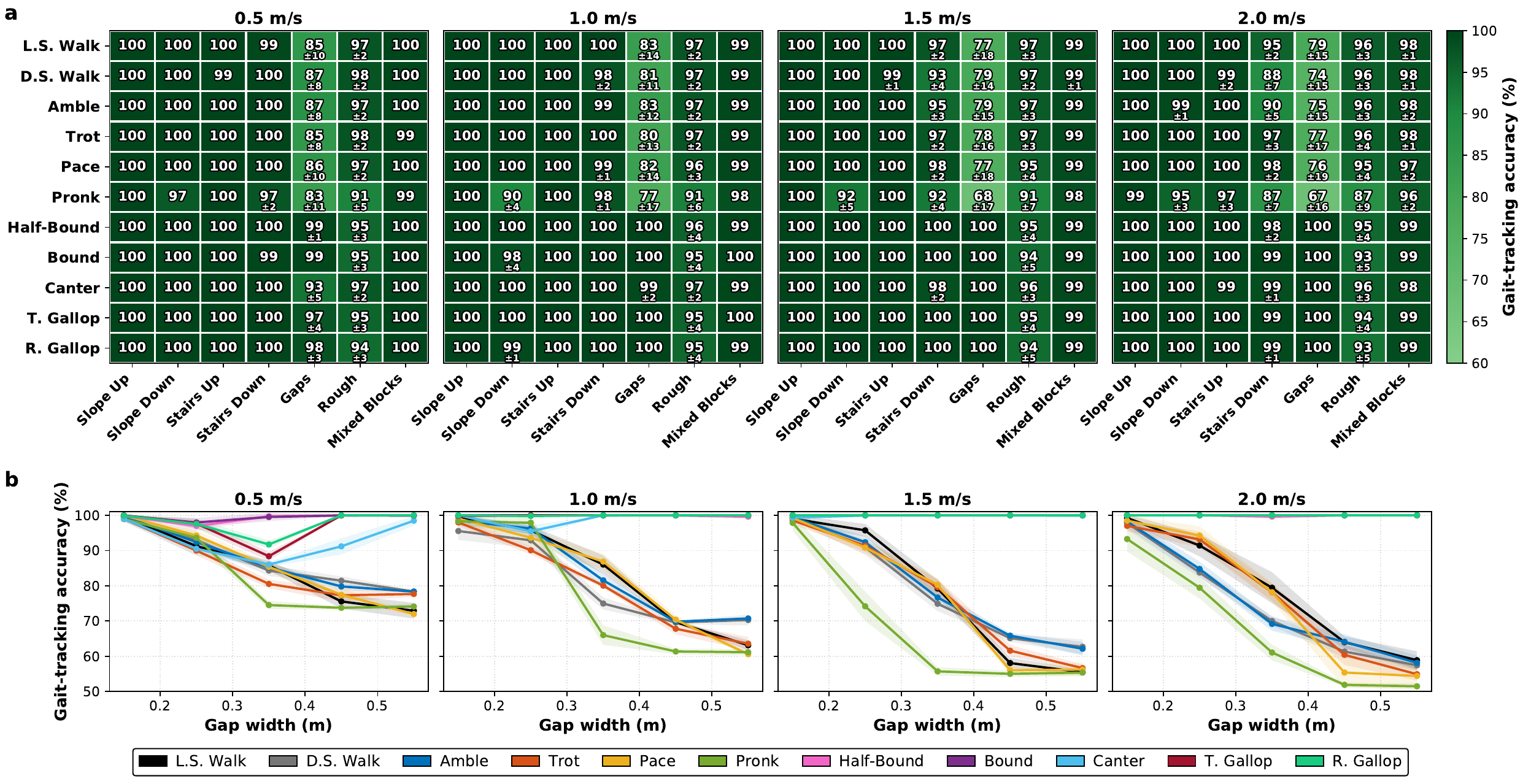}
    \caption{\textbf{Realization of commanded gait patterns across terrains and speeds}. Gait-tracking accuracy measures the fraction of commanded stance time during which the corresponding foot is in terrain contact. We evaluate 11 gaits across seven terrains, five difficulty levels, and four commanded velocities (0.5–2.0 m/s), with 100 robots per setting. (a) Mean gait-tracking accuracy for each gait, terrain, and velocity. Values are averaged across difficulty levels, with standard deviations indicated where applicable. Commanded stance phases are consistently realized on most terrains, typically with 95–100\% accuracy, with the largest reductions occurring over gaps. (b) Gait-tracking accuracy as a function of gap width. Walking, ambling, trotting, pacing, and pronking progressively lose contact availability as gaps widen, whereas bounding and galloping gaits maintain near-perfect contact realization. These reductions are primarily associated with the absence of suitable footholds over gaps, rather than changes in the commanded CPG phase pattern.}
    \label{fig:supp_gait_tracking}
\end{figure}

\section{Stage~1 Comprehensive Metrics}

In Fig.~3 (main text), we presented the gaits that minimize three key metrics: cost of transport, mean angular velocity, and mean joint acceleration. Here we also show the per-gait trends and comparisons across the same velocity--terrain combinations. Additionally, noting that other quantities may be equally, or even more, important, depending on the task or desired behavior, we show additional metrics including the traversal success rate, mean CPG amplitude (related to step length), mean CPG frequency (approximate step frequency), mean body height, mean swing foot ground clearance, and mean foot $x$ offset. Any of these quantities could also be used to optimize in Stage~2, for example.

In Supplementary Figures 3-11, we plot metrics vs.\ four velocities and five terrain difficulties averaged across 100 agents in parallel, as in the main text. For a given velocity, the mean performance across all terrain difficulties is denoted with a circle, and the error bars show the standard deviation across the full spectrum of terrain difficulties from 1 to 5. If applicable, the optimal average gait for a given terrain-velocity across all difficulties is denoted with a star (*). We also compare a baseline policy trained with the exact same terrain scenario and reward function, but \textit{without} coupling (\ie\ $w_{ij}=0$), which reduces Equation~2 (main text) to $\dot{\theta}_i = \omega_i$. This baseline can freely coordinate each of its limbs to form its own gaits and gait transitions.

Supplementary Fig.~3 shows the traversal success rate for each gait across all seven terrains. Most gaits achieve a near perfect traversal success rate across slopes, stairs, gaps, and rough terrain, although there are some reductions on stairs, and the pronk achieves 90--95\% success rate on rough terrain. The mixed blocks terrain is more challenging, with slightly lower success rates and larger differences between gaits. Falls here are most often due to not facing an obstacle straight on, so for example having a large box under the feet on the left side, but not the right side. In these cases, the policy can do little to recover as we have framed the locomotion problem primarily in the sagittal plane. On these varied parkour-like terrains, the bound-style gaits like half-bound, bound, and the gallops are best suited to traverse without falling. The baseline performs roughly as well as the transverse gallop, and in fact has a similar gait pattern.

Supplementary Fig.~4 shows the mean COT for each of the gaits on the different terrains. Generally, the traditionally `slower' gaits like the walks, amble, trot, and pace are more energy efficient when the terrain is not too difficult. However, for very fast and dynamic terrain traversal like ascending stairs at high velocity or crossing gaps, the more aggressive bound-style gaits are more energy efficient. Notably, although we might expect the baseline without gait coupling to be able to adapt to any terrain more efficiently as it is not constrained to perform a particular gait, we find that it is not optimal in most cases. This highlights that choosing the right gait is critical to optimizing performance, and further showcases that gaits that emerge when learning in general can be local optima.

Supplementary Fig.~5 shows the mean base angular velocity for each of the gaits on the different terrains. As could be expected, due to symmetry, the trot and bound gaits effectively minimize this quantity best across almost all of the terrains. Notably, the baseline has learned a policy that prioritizes other metrics, as there are only a couple instances where it minimizes the base angular velocity better than the fixed gaits.

Supplementary Fig.~6 shows the mean total joint acceleration for each of the gaits on the different terrains. The differences are more subtle for this metric for most terrain scenarios, with some exceptions. For the gaps, the bound-type gaits have much lower acceleration, as they do not need to quickly cycle through steps. Notably, the baseline has the lowest joint acceleration in most of the considered velocity--terrain scenarios, showing that the agent may have optimized this quantity over other metrics like mean angular velocity or COT. 

Supplementary Figs.~7 and~8 show the mean oscillator amplitudes and frequencies, respectively, for each of the gaits on the different terrains. Observing these two quantities together gives insights into how the agent coordinates gait parameters based on the terrain. For example, for the gaps, the bound-style gaits use higher amplitudes (corresponding to larger step lengths) with lower step frequencies, compared to the walk-style gaits. All gaits use smaller steps when descending slopes, and the bound-style gaits use faster/smaller steps when descending stairs. Notably, the baseline consistently locomotes with larger step lengths and lower step frequencies across all terrains.

Supplementary Fig.~9 shows the mean body height for each of the gaits on the different terrains. Some general trends include decreasing the base height with faster velocity on stairs/slopes up. There are also some noticeable clusters among the walk-style and bound-style gaits on different terrains: bound-style has higher body height on slopes down and stairs up/down, while walk-style maintains a higher body height on the rest. The pronk gait is an exception, as it maintains a lower body height on most terrains.
The baseline greatly changes the height strategy when crossing the gaps. Normally, it locomotes with one of the highest body heights, except for on the gaps, where this trend reverses.

Supplementary Fig.~10 shows the mean swing foot ground clearance for each of the gaits on the different terrains. Generally, the ground clearance increases with the locomotion velocity, and increases more significantly for the walk-style gaits. The pronk gait uses a substantially lower ground clearance than the other gaits on all terrains.

Supplementary Fig.~11 shows the mean foot $x$-offset for each of the gaits on the different terrains. This gives insight into adapting the posture based on the terrain. For example, for descending slopes or stairs, the foot offset needs to be more in front to avoid falling, while the opposite is true for ascending stairs/slopes. The baseline typically keeps the foot offset a lot closer to 0 (under the hip) than the fixed gaits on most of the terrains.

\newpage 
\begin{figure}[H]
    \vspace{-1em}
    \centering
    \suppfigtoc{fig:supp_success}
        {Traversal success rate across terrains}
     \includegraphics[width=\textwidth]{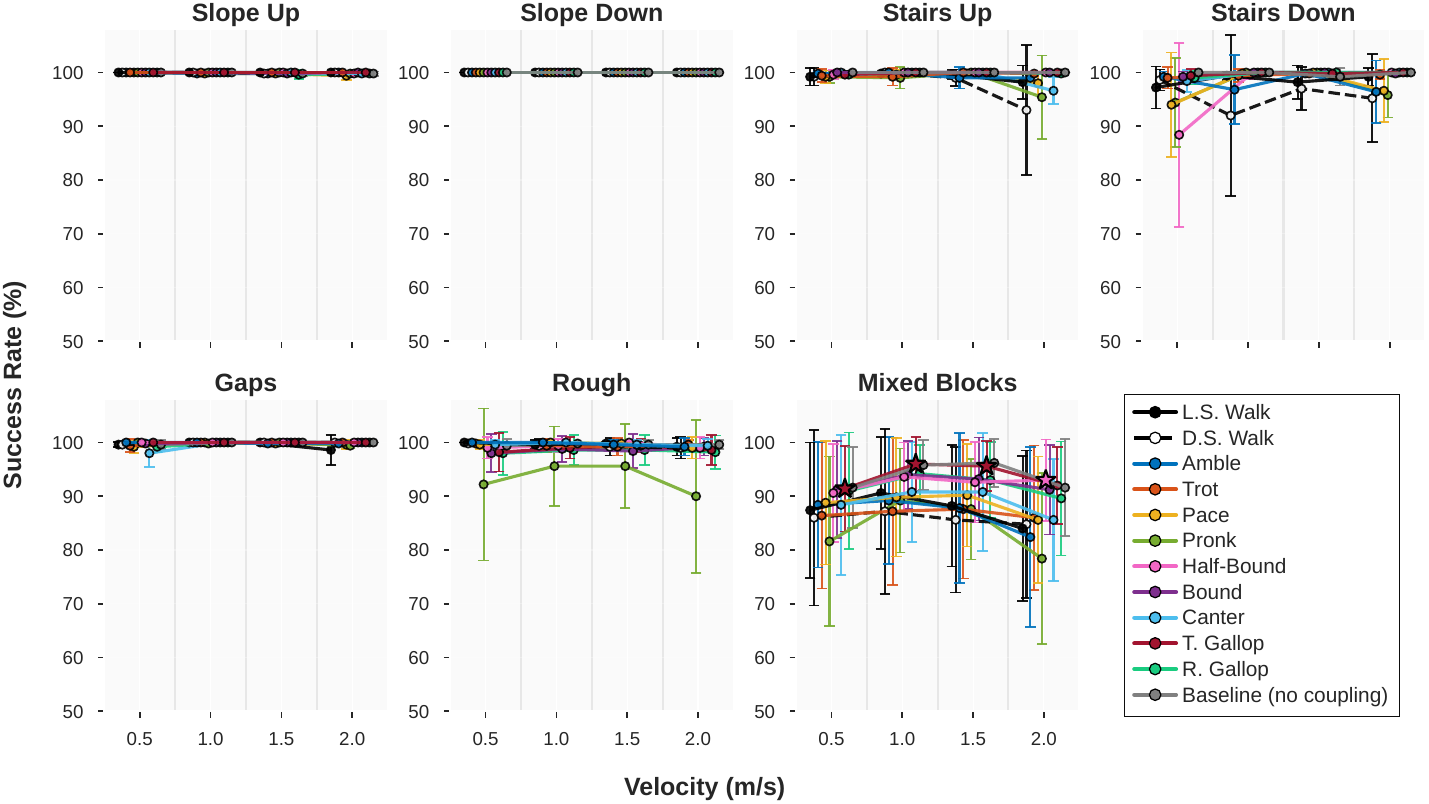}\\
    \vspace{-0.5em}
    \caption{Traversal success rate for all gaits across all terrains. A baseline trained without gait coupling is shown in gray. The mean across all terrain difficulties is shown with the filled circle, and the error bars denote the standard deviation across the terrain difficulties from 1 through 5. If not a tie, the optimal gait from among the 11 is denoted by a star (*) for each velocity. Notably, even though the baseline does not have a fixed gait, implying it could adapt better to terrains in general, it performs comparably to the transverse gallop.
    }
    \vspace{-1em}
    \label{fig:supp_success}
\end{figure}

\begin{figure}[H]
    \centering
    \suppfigtoc{fig:supp_cot}
        {Cost of Transport across terrains}
    \includegraphics[width=\textwidth]{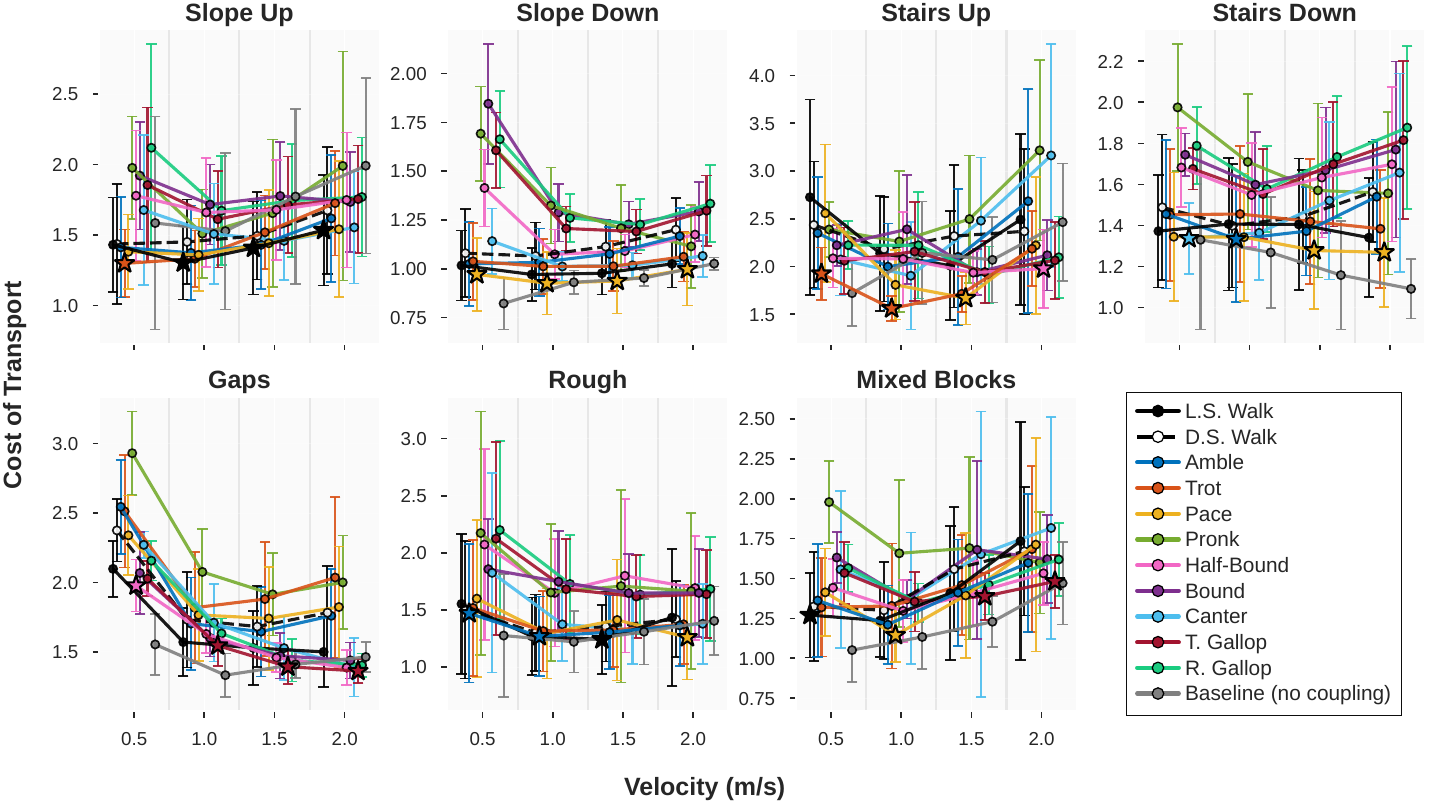}\\
    \vspace{-0.5em}
    \caption{Cost of Transport for all gaits across all terrains. A baseline trained without gait coupling is shown in gray.
    The mean across all terrain difficulties is shown with the filled circle, and the error bars denote the standard deviation across the terrain difficulties from 1 through 5.
    The optimal gait from among the 11 is denoted by a star (*) for each velocity. Notably, even though the baseline does not have a fixed gait, implying it could adapt better to terrains in general, there are few terrain scenarios where it outperforms the optimal gait.
    }
    \vspace{-1em}
    \label{fig:supp_cot}
\end{figure}

\begin{figure}[H]
    \centering
    \suppfigtoc{fig:supp_ang}
        {Mean base angular velocity across terrains}
    \includegraphics[width=\textwidth]{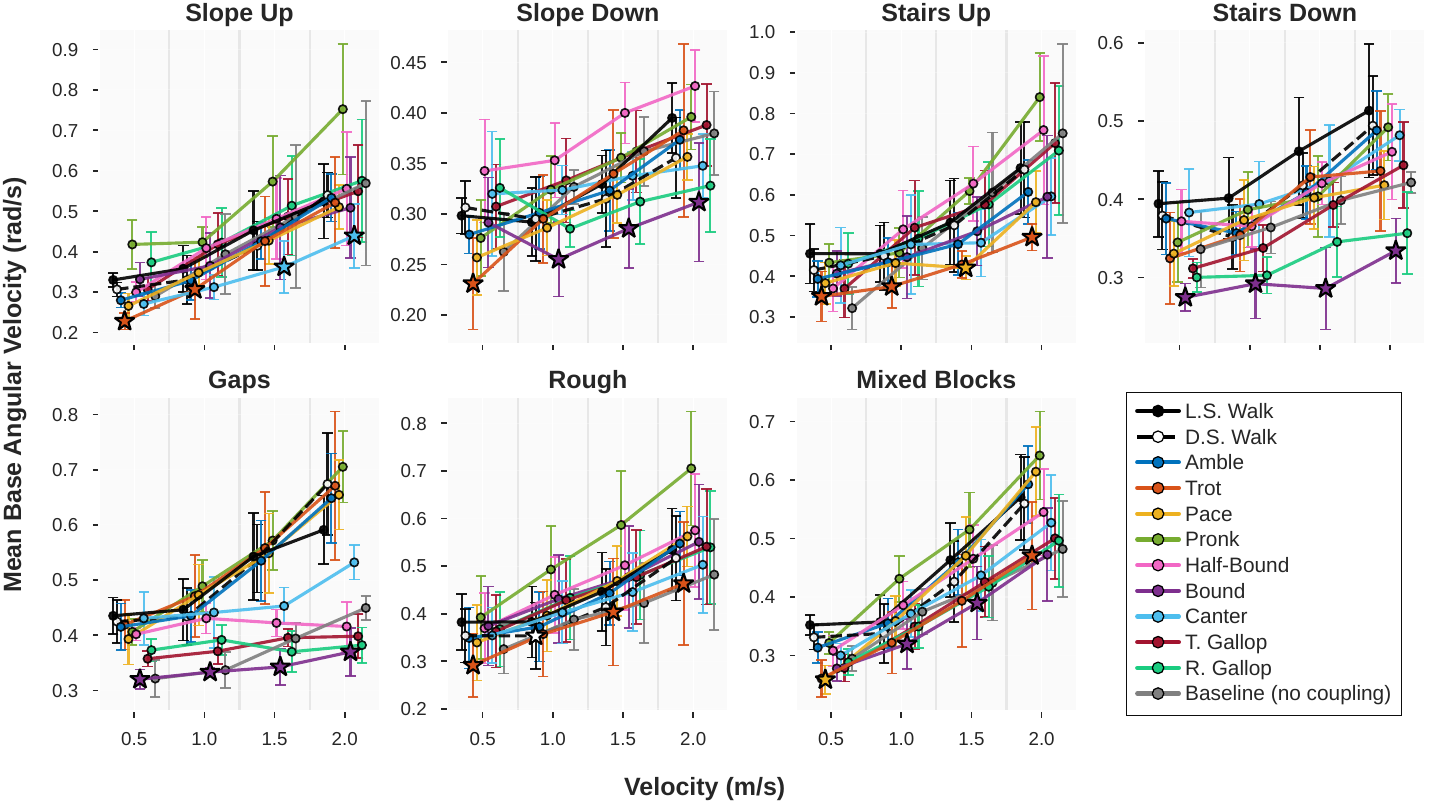}\\
    \vspace{-0.5em}
    \caption{Mean base angular velocity for all gaits across all terrains. A baseline trained without gait coupling is shown in gray.
    The mean across all terrain difficulties is shown with the filled circle, and the error bars denote the standard deviation across the terrain difficulties from 1 through 5.
    The optimal gait from among the 11 is denoted by a star (*) for each velocity.
    }
    \label{fig:supp_ang}
\end{figure}

\begin{figure}[H]
    \centering
    \suppfigtoc{fig:supp_acc}
        {Mean total joint acceleration across terrains}
    \includegraphics[width=\textwidth]{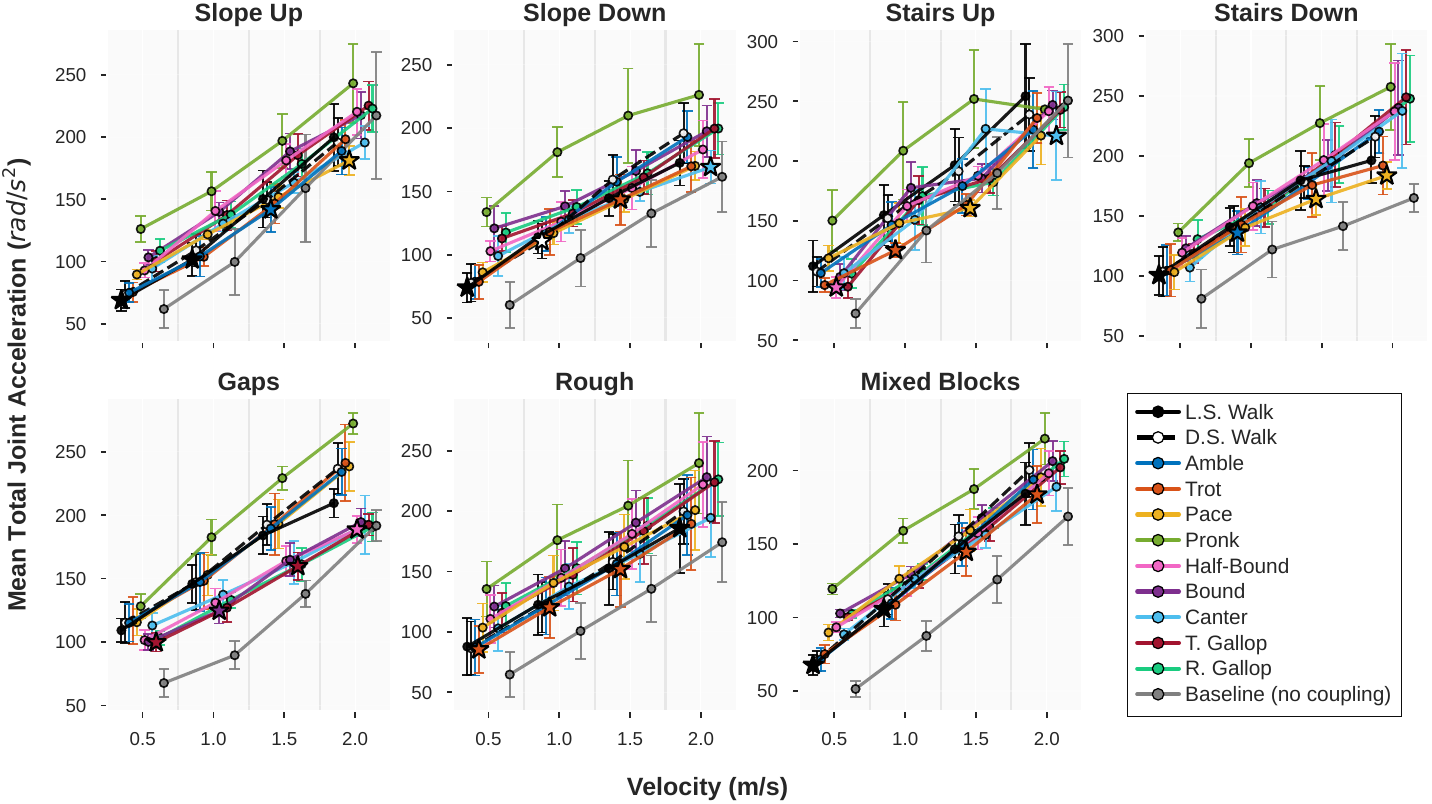}\\
    \vspace{-0.5em}
    \caption{Mean total joint acceleration for all gaits across all terrains. A baseline trained without gait coupling is shown in gray.
    The mean across all terrain difficulties is shown with the filled circle, and the error bars denote the standard deviation across the terrain difficulties from 1 through 5.
    The optimal gait from among the 11 is denoted by a star (*) for each velocity.
    }
    \label{fig:supp_acc}
\end{figure}

\begin{figure}[H]
    \centering
    \suppfigtoc{fig:supp_cpg_amp}
        {Mean oscillator amplitudes (related to step lengths) across terrains}
    \includegraphics[width=\textwidth]{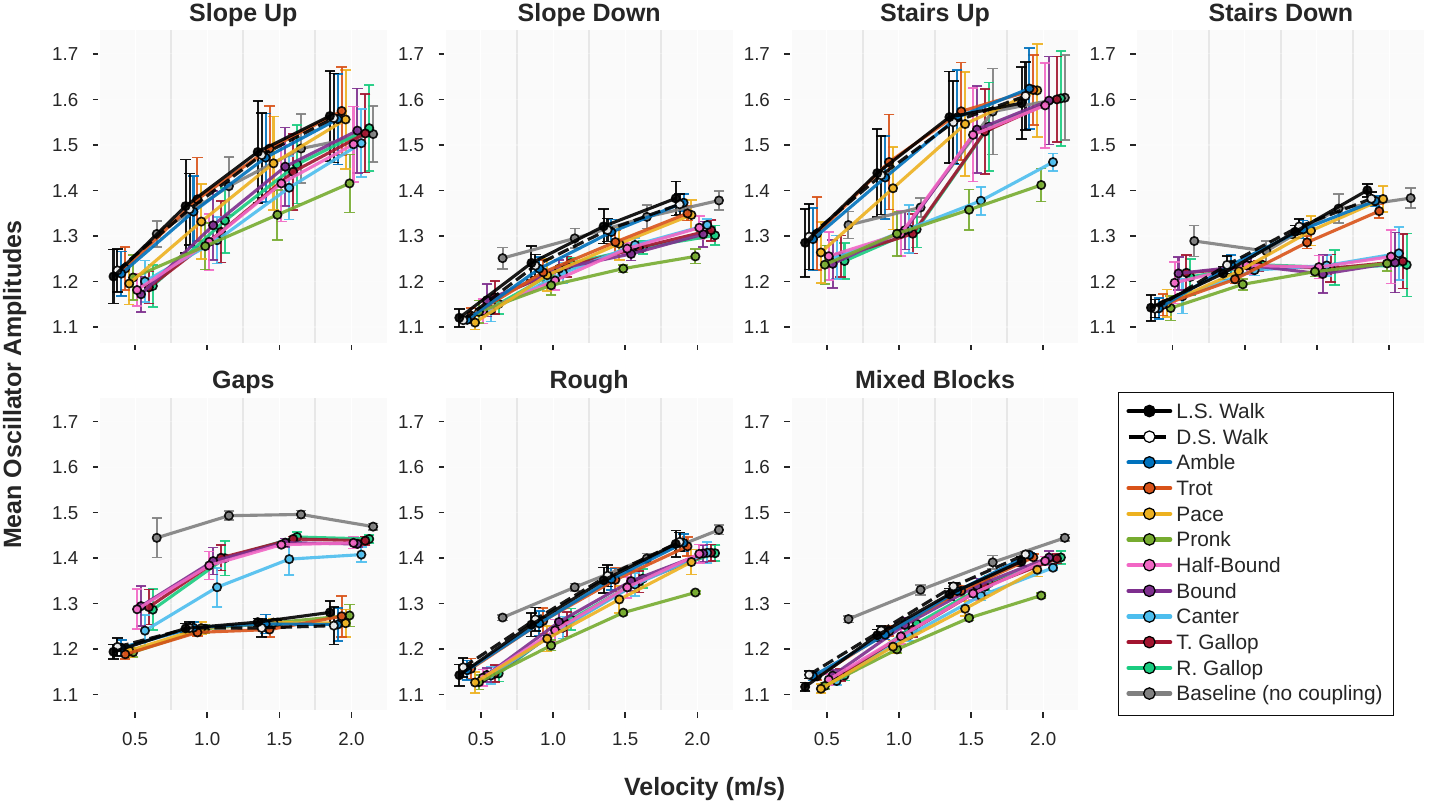}\\
    \vspace{-0.5em}
    \caption{Mean oscillator amplitudes (related to step lengths) for all gaits across all terrains. A baseline trained without gait coupling is shown in gray.
    The mean across all terrain difficulties is shown with the filled circle, and the error bars denote the standard deviation across the terrain difficulties from 1 through 5.
    }
    \label{fig:supp_cpg_amp}
\end{figure}

\begin{figure}[H]
    \centering
    \suppfigtoc{fig:supp_cpg_freq}
        {Mean oscillator frequencies (approximate step frequencies) across terrains}
    \includegraphics[width=\textwidth]{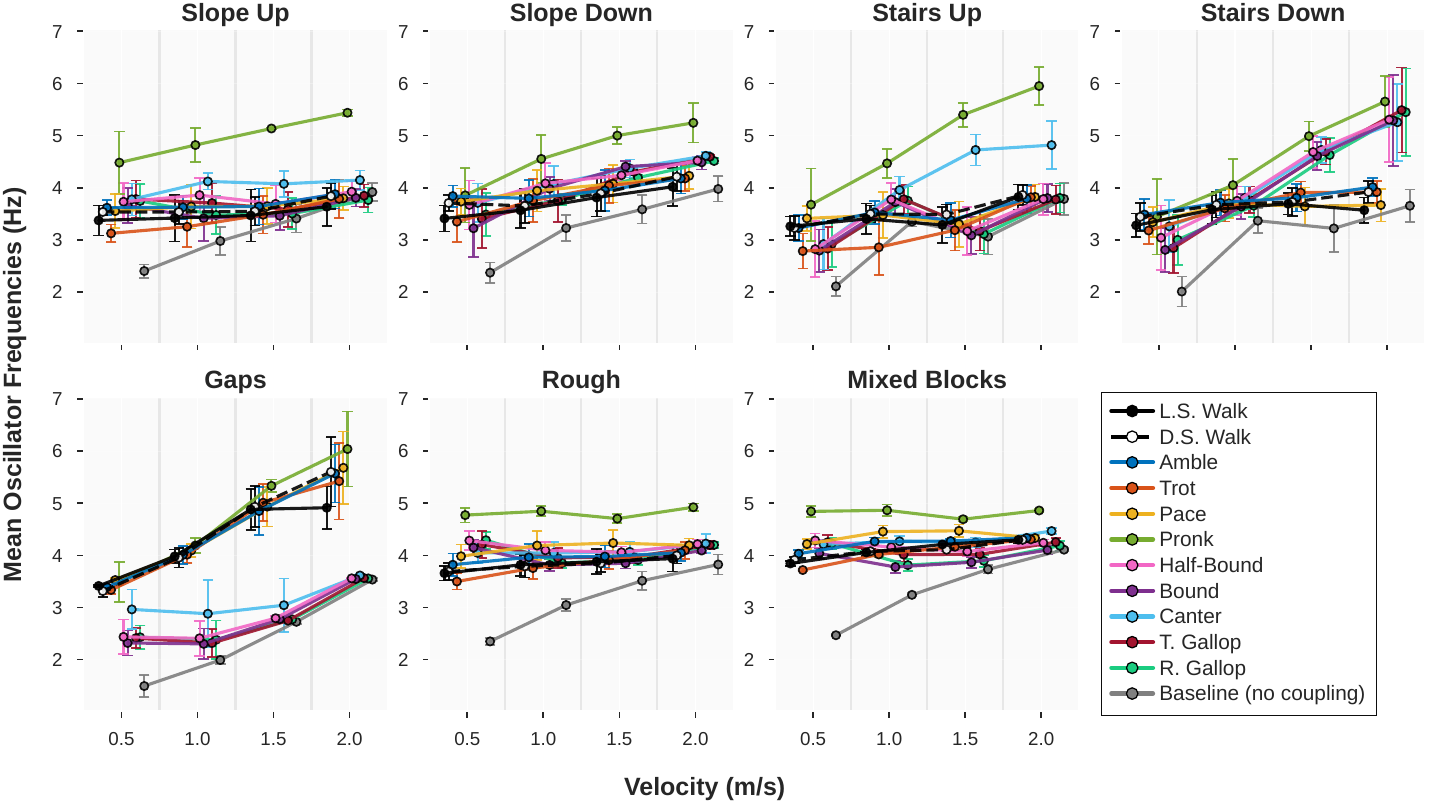}\\
    \vspace{-0.5em}
    \caption{Mean oscillator frequencies (approximate step frequencies) for all gaits across all terrains. A baseline trained without gait coupling is shown in gray.
    The mean across all terrain difficulties is shown with the filled circle, and the error bars denote the standard deviation across the terrain difficulties from 1 through 5.
    }
    \label{fig:supp_cpg_freq}
    \vspace{-0.5em}
\end{figure}

\begin{figure}[H]
    \centering
    \suppfigtoc{fig:supp_bh}
        {Mean body height across terrains}
    \includegraphics[width=\textwidth]{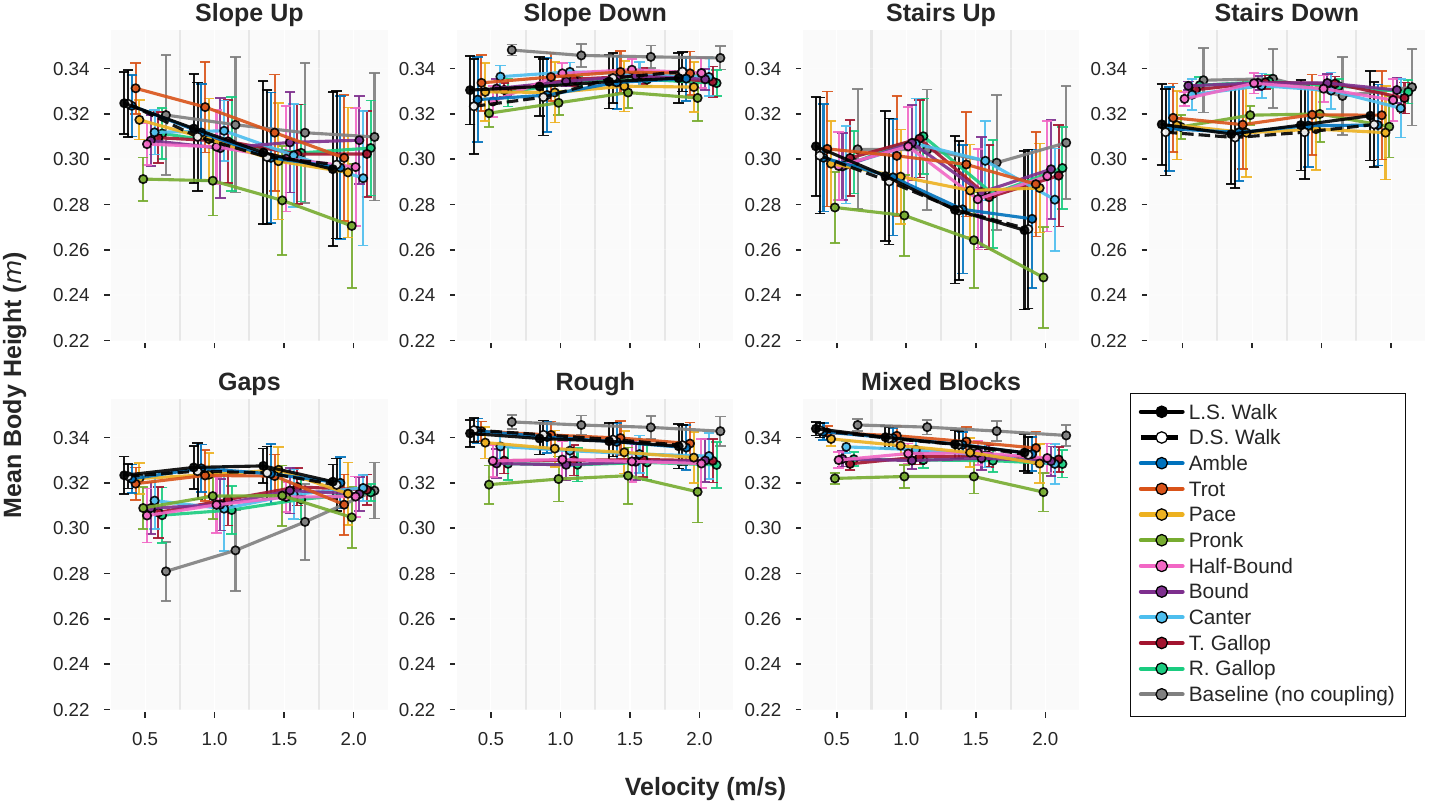}\\
    \vspace{-0.5em}
    \caption{Mean body height for all gaits across all terrains. A baseline trained without gait coupling is shown in gray.
    The mean across all terrain difficulties is shown with the filled circle, and the error bars denote the standard deviation across the terrain difficulties from 1 through 5.
    }
    \label{fig:supp_bh}
\end{figure}

\begin{figure}[H]
    \centering
    \suppfigtoc{fig:supp_gc}
        {Mean swing foot ground clearance across terrains}
    \includegraphics[width=\textwidth]{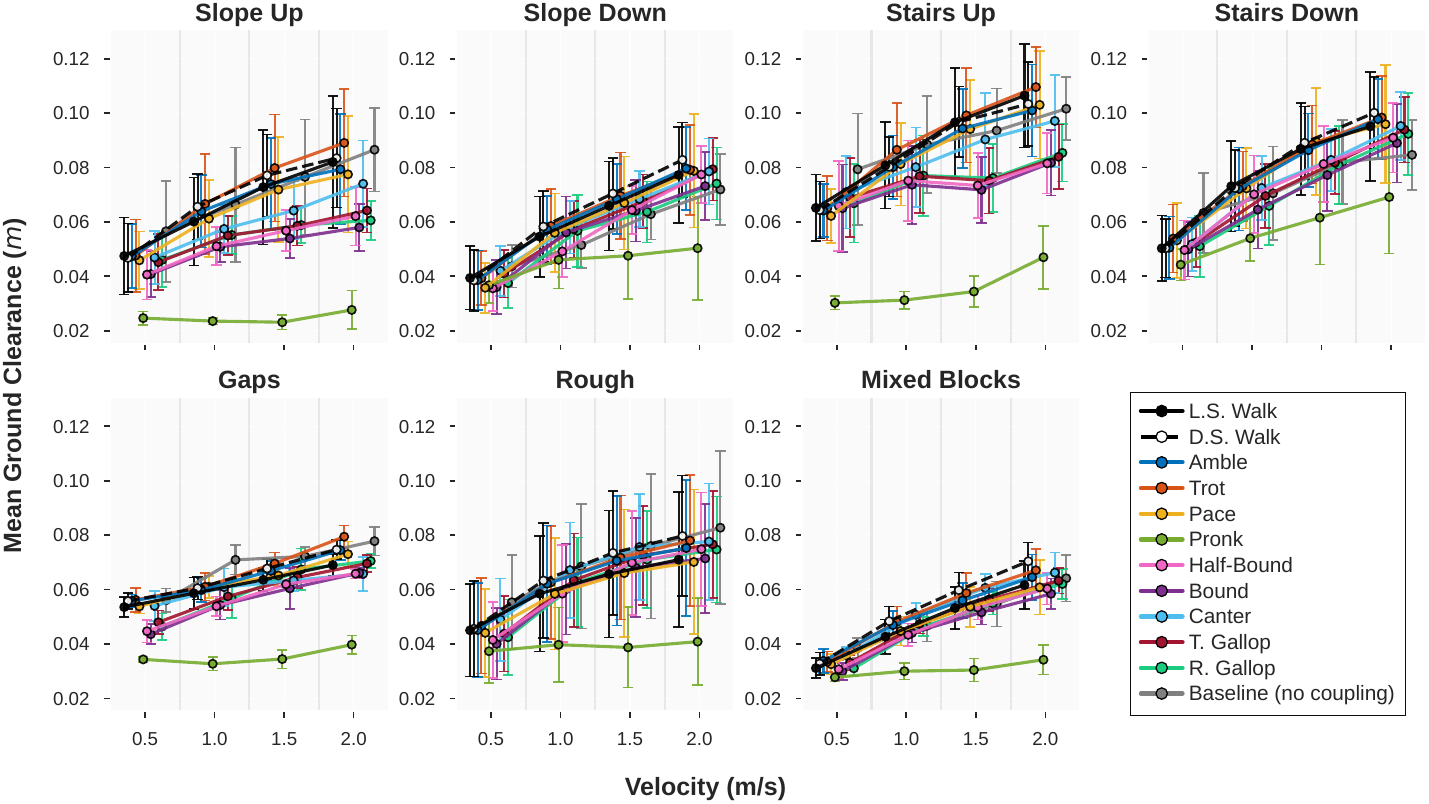}\\
    \vspace{-0.5em}
    \caption{Mean swing foot ground clearance for all gaits across all terrains. A baseline trained without gait coupling is shown in gray.
    The mean across all terrain difficulties is shown with the filled circle, and the error bars denote the standard deviation across the terrain difficulties from 1 through 5.
    }
    \label{fig:supp_gc}
\end{figure}

\begin{figure}[H]
    \centering
    \suppfigtoc{fig:supp_xoff}
        {Mean foot \texorpdfstring{$x$}{x}-offset across terrains}
    \includegraphics[width=\textwidth]{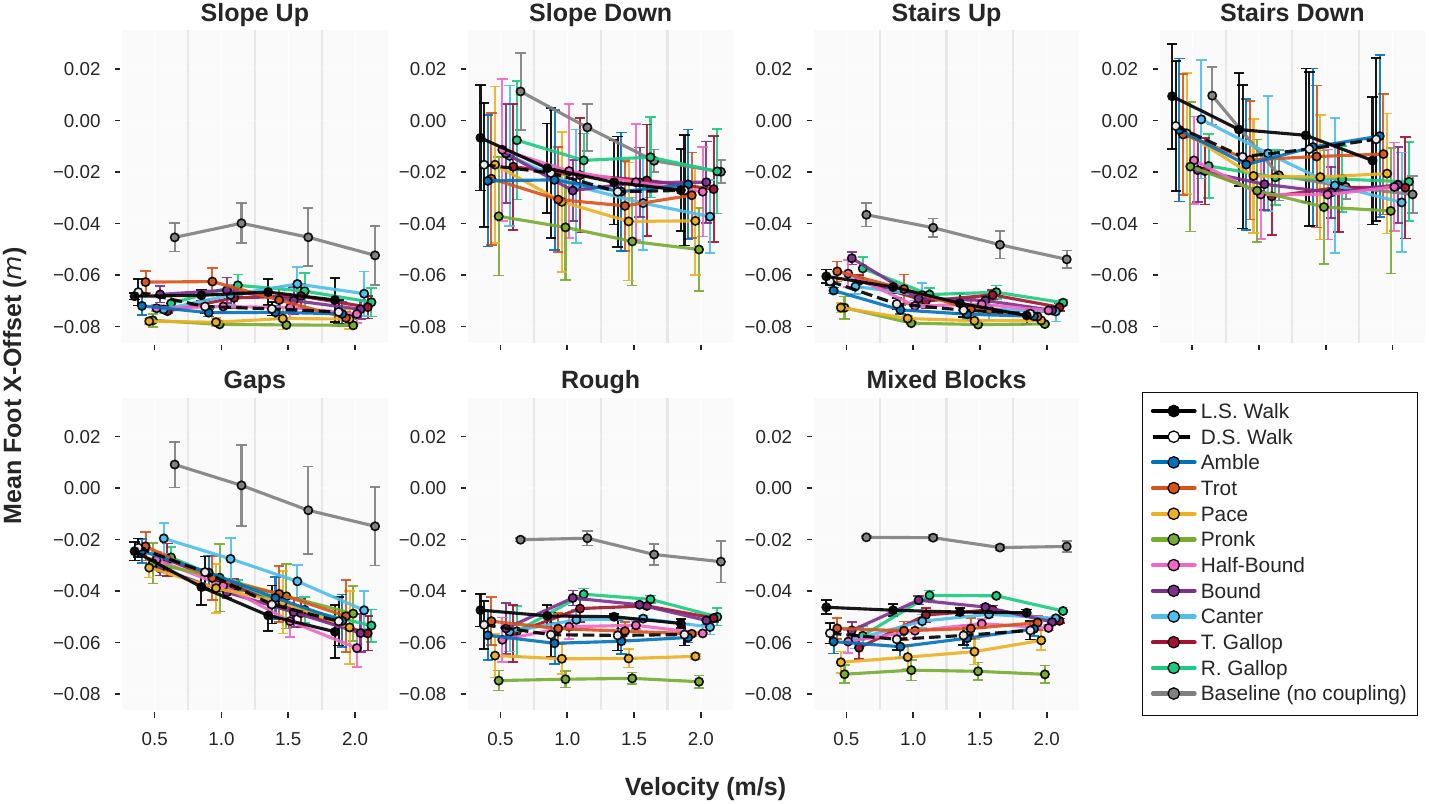}\\
    \vspace{-0.5em}
    \caption{Mean foot $x$-offset for all gaits across all terrains. A baseline trained without gait coupling is shown in gray.
    The mean across all terrain difficulties is shown with the filled circle, and the error bars denote the standard deviation across the terrain difficulties from 1 through 5.
    }
    \label{fig:supp_xoff}
\end{figure}

\newpage
\section{Stage~2 Policies with Diverse Reward Functions}

We train multiple individual Stage~2 policies with different performance criteria, as shown in Supplementary Fig.~12. The expressions shown define the criterion-specific reward terms. All Stage~2 policies additionally use an action-rate penalty and an out-of-range action penalty (rewards 5 and 6 of Supplementary Table 2), with the former discouraging frequent gait switching. Above the red dashed line, the policies optimize conventional performance criteria, minimizing quantities such as power or angular velocity, or maximizing velocity-command tracking. Below the red line, we instead maximize quantities normally minimized to showcase the expressivity and diverse behaviors possible.

For each policy, we test 100 agents in parallel at four different velocities on the same parkour environment from Fig.~4 (main text), and observe the percentage of time each gait is chosen. Above the red line, the bound gait is frequently selected for minimizing quantities that do not include the pitch axis, consistent with its symmetry. Also interesting are the gaits that instead minimize pitch rate, for which the bound still appears as the best gait some percentage of the time. The agent also finds that the bound gait effectively minimizes contact forces relative to other gaits in our training scheme, but is not the best for velocity tracking.

Below the red line, we consider gait selections that instead maximize the quantities that we would normally want to minimize. This shows which gaits perform worst under these criteria, while demonstrating the versatility and expressivity of the framework. As can be expected, the pronk gait appears frequently across these tests, as it has difficulty keeping a flat base on challenging terrain, and uses a lot of joint acceleration to jump. When maximizing power, the policy learns to switch between gaits more frequently, creating large phase jumps that increase power consumption, and often prefers the rotary or transverse gallops.

\begin{figure}[H]
    \centering
    \suppfigtoc{fig:supp_stage2_all}
        {Gait selection to optimize diverse performance objectives}
    \includegraphics[width=\textwidth]{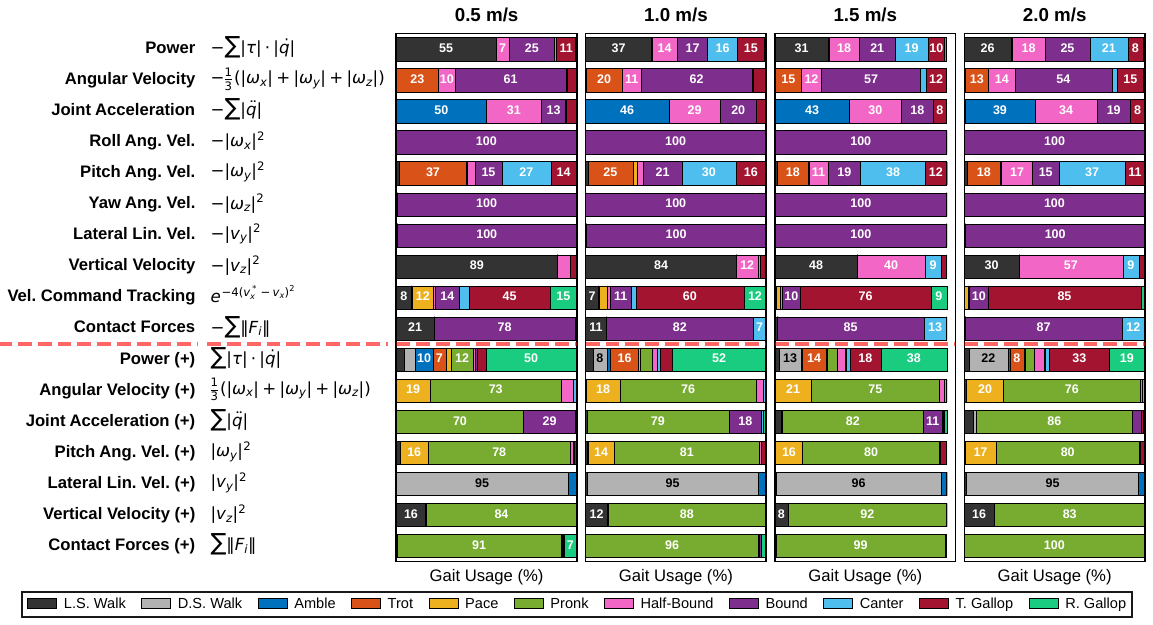}\\
    \vspace{-0.6em}
    \caption{
    Gait selection under different Stage 2 performance criteria. Independently trained Stage 2 policies are evaluated on the parkour terrain from Fig.~4 (main text). The percentage of time each gait is selected is shown across four commanded velocities. The expressions indicate the criterion-specific reward terms, with shared action regularization terms. Above the dashed red line, policies minimize quantities such as power, angular velocity, or contact forces. Below the red line, policies instead maximize some of these quantities to demonstrate the diversity of gait-selection behaviors, for example for the least energy-efficient locomotion possible, or the greatest vertical oscillations. For minimizing quantities that do not include the pitch axis, the bound gait is frequently selected due to its symmetry. When maximizing several of these quantities, the pronk gait is often preferred.
    }
    \label{fig:supp_stage2_all}
    \vspace{-1em}
\end{figure}

\newpage
\section{Training details and comparisons}

\begin{table}[H]
\centering
\caption{PPO hyperparameters and neural network architectures.}
\label{tab:supp_ppo}
\supptabtoc{PPO hyperparameters and
network architecture}
\begin{tabular}{lc}
\hline
Parameter & Value \\
\hline
Batch size & 98,304 (4,096 $\times$ 24) \\
Mini-batch size & 24,576 (4,096 $\times$ 6) \\
Number of epochs & 5 \\
Clip range & 0.2 \\
Entropy coefficient & 0.01 \\
Discount factor & 0.99 \\
GAE discount factor & 0.95 \\
Desired KL-divergence $kl^*$ & 0.01 \\
Learning rate & Adaptive \\
Hidden layers (Stage~1 and Stage~2 networks) & 3 \\
Hidden units per layer (Stage~1 and Stage~2 networks) & [512, 256, 128] \\
Activation & ELU \\
\hline
\end{tabular}
\end{table}

\begin{table}[H]
\centering
\caption{\textbf{Stage~1 reward function terms}. The same reward is used for all 11 gaits and all terrains, with no term referring to the gait, the body height, the foot ground clearance, or the foot offset, all of which the policy controls directly through the CPG interface. $(\cdot)^{*}$ denotes a desired command and $f(x) := \exp(-\|x\|^2/0.25)$. $dt = 0.03$\,s is the control time step. 
}
\label{tab:supp_reward}
\supptabtoc{Stage~1 reward function terms}
\begin{tabular}{llc}
\hline
Term & Formula & Weight \\
\hline
Linear velocity tracking $v_{b,x}^{*}$ & $f(v_{b,x}^{*} - v_{b,x})$ & $3\,dt$ \\
Lateral/vertical velocity $\bm{v}_{b,yz}$ & $-\|\bm{v}_{b,yz}\|^2$ & $2\,dt$ \\
Angular velocity $\bm{\Omega}_{b,xyz}$ & $-\|\bm{\Omega}_{b,xyz}\|^2$ & $0.1\,dt$ \\
Power & $-\sum_i |\tau_i\dot{q}_i|$ & $0.001\,dt$ \\
Action rate & $-\|\bm{a}_{t-1}-\bm{a}_t\|^2$ & $0.01\,dt$ \\
Out-of-range actions & $-\sum_k \mathbb{I}(|a^{\mathrm{raw}}_{t,k}|> a_{\max})$ & $1\,dt$ \\
Straight-line alignment & $-\psi_{b,y}^2-(y_b-y_0)^2$ & $3\,dt$ \\
\hline
\end{tabular}
\end{table}

\vspace{3em}
\subsection{Supplementary Table 3: Comparison with learning-based parkour controllers}

Supplementary Table~3 compares our framework with representative learning-based parkour controllers. These methods address related but distinct research questions: previous approaches primarily focus on agile obstacle traversal and navigation, whereas our work investigates the execution, evaluation, and selection of a broad repertoire of quadrupedal gaits with terrain-adaptive locomotion styles. The comparison highlights differences in reported capabilities and training requirements, rather than providing a head-to-head performance benchmark. Our framework differs by generating and systematically evaluating the performance of individually commandable gait couplings across terrains, commanded velocities, and performance metrics (Fig. 3, main text).

\begin{table}[H]
\centering
\caption{Reported capabilities and training organization of representative learning-based parkour controllers and our gait-learning framework. The methods address different research questions; repertoire counts refer to parkour skills, locomotion modes, or gait couplings, and are not directly equivalent.}
\label{tab:supp_comparison}
\begingroup
\fontsize{8.15}{10.0}\selectfont
\setlength{\tabcolsep}{2.5pt}
\renewcommand{\arraystretch}{1.16}
\setlength{\extrarowheight}{0.9pt}
\renewcommand{\tabularxcolumn}[1]{m{#1}}
\newcommand{\tyes}{$\checkmark$}
\newcommand{\tnr}{--}
\begin{tabularx}{\textwidth}{@{}>{\raggedright\arraybackslash}m{0.3\textwidth}*{5}{>{\centering\arraybackslash}X}@{}}
\toprule
 & \shortstack{\textbf{Extreme}\\\textbf{Parkour}\cite{cheng2024extreme}}
 & \shortstack{\textbf{Robot}\\\textbf{Parkour}\\\textbf{Learning}\cite{zhuang2023robot}}
 & \shortstack{\textbf{ANYmal}\\\textbf{Parkour}\cite{hoeller2024anymal}}
 & \shortstack{\textbf{Agile}\\\textbf{Jumping}\cite{yang2024agile}}
 & \shortstack{\textbf{This}\\\textbf{work}} \\
\midrule
\multicolumn{6}{@{}l}{\textbf{Gait repertoire, control, and evaluation}}\\[2pt]
Gait repertoire and command interface
 & 2 modes & 5 skills & 5 skills & Fixed bounding 
 & \shortstack{\textbf{11 gaits +}\\
  \textbf{zero-shot gait}\\
  \textbf{generalization}} \\
Systematic gait comparison across terrains, commanded speeds, and metrics
 & \tnr & \tnr & \tnr & \tnr & \tyes \\
 Velocity tracking over varying commanded speeds
 & \tnr & \tnr & \tnr & \tnr & \tyes \\
Commandable locomotion style and extrapolation (\ie~height, clearance)
 & \tnr & \tnr & \tnr & \tnr & \tyes \\
 \addlinespace[3pt]
 \shortstack[l]{Interchangeable objectives\\(gait-selection criteria)}
 & \tnr & \tnr & \tnr & \tnr & \tyes \\
Bipedal-mode locomotion
 & \shortstack{front legs} & \tnr & \tnr & \tnr
 & \shortstack{front+rear legs} \\
\addlinespace[3pt]
\midrule
\multicolumn{6}{@{}l}{\textbf{Parkour and deployment}}\\[3pt]
Onboard terrain perception and physical-robot validation
 & \tyes & \tyes & \tyes & \tyes & \tyes \\
Non-rhythmic obstacle-specific maneuvers 
 & \tyes & \tyes & \tyes & \tnr & \tnr \\
Terrain-adaptive direction changes
 & \tyes$^{a}$ & \tnr & \tyes & \tnr & \tnr \\
Quantified success beyond training terrain difficulty
 & \tyes$^{b}$ & \tnr & \tyes$^{b}$ & \tnr & \tyes$^{c}$ \\
Transfer across robot models without retraining the low-level policy
 & \tnr & \tnr & \tnr & \tnr & \tyes$^{d}$ \\
\addlinespace[3pt]
\midrule
\multicolumn{6}{@{}l}{\textbf{Training}}\\[3pt]
Locomotion policy training
 & \shortstack{RL\\(oracle heading)}
 & \shortstack{5 skill\\ policies (RL)}
 & \shortstack{5 skill\\ policies (RL)}
 & \shortstack{Centroidal\\motion RL}
 & \shortstack{Stage 1 RL\\ (11 gaits)} \\
\addlinespace[5pt]
Additional learned components
 & \shortstack{Vision+heading\\distillation}
 & \shortstack{Multi-skill\\ vision\\distillation}
 & \shortstack{Navigation RL\\$+$ perception}
 & \shortstack{Heightmap\\predictor}
 & \shortstack{Stage 2 RL selector\\+ visual encoder} \\

 \addlinespace[3pt]
Reward terms (approx.)$^{e}$
 & 14 & Up to 14 & 16 & 12 & \textbf{7} \\
 \addlinespace[5pt]
Reward reuse across behaviors
 & \shortstack{Shared core;\\handstand term}
 & \shortstack{Some shared;\\varying weights}
 & \shortstack{Varies by\\skill}
 & \shortstack{One jumping\\behavior}
 & \shortstack{Identical across\\11 gaits$^{f}$} \\
\bottomrule
\end{tabularx}
\par\vspace{7pt}
\begin{minipage}{\textwidth}
\fontsize{7.55}{9.35}\selectfont
\textbf{Notes.} \tyes, reported or demonstrated; --, not established in the cited work. A \emph{gait} is an interlimb phase/contact pattern, whereas a parkour skill can involve non-rhythmic maneuvers; repertoire counts are therefore not directly comparable. Our 11 trained gait couplings are individually commandable, and the same policy also executes previously unseen gait couplings zero-shot. Locomotion style (body height, swing foot clearance, and foot offset) can be user-adjusted, including overrides beyond training ranges.

\textsuperscript{a}Extreme Parkour learns heading adjustments from waypoint supervision. ANYmal Parkour additionally employs high-level navigation and skill selection; our Stage~2 policies select gaits, not routes. 

\textsuperscript{b}Extreme Parkour demonstrates successful traversal of 0.50\,m steps (Fig.~7 of that work), exceeding the nominal 0.45\,m maximum specified in its released training configuration. ANYmal Parkour evaluates success rates up to 120\% of the maximum training difficulty (Fig.~4F of that work), with successful extrapolation most evident for walking.

\noindent\textsuperscript{c}For our framework, the tested gaps are 45\% wider, stairs 60\% higher, and walls 10\% higher than training maxima. Different studies use different generalization protocols. 

\textsuperscript{d}Our low-level networks are deployed on Go1 and Go2 hardware, with additional simulated embodiments.

\textsuperscript{e}Reward counts (14, up to 14, 16, 12, and 7) are indicative weighted-component counts from reported formulations and implementations; regularization/shaping and termination conventions differ and the external counts are not standardized. 

\textsuperscript{f}The same seven Stage~1 terms and weights are used for all 11 gaits and seven terrains, without gait- or style-tracking rewards (Supplementary Table~2). Stage~2 selectors can be retrained for different criteria while Stage~1 remains frozen.
\end{minipage}
\endgroup
\end{table}

\newpage 
\section{Supplementary Videos}
\subsection{Supplementary Video 1}

A combined simulation video demonstrating: (1)~all 11 gaits on all terrains from Stage~1, (2)~Stage~2 gait selection and transitions on the parkour terrain, (3)~zero-shot generalization to novel gaits, parameter overrides, out-of-distribution terrains, and (4)~bipedal-mode (handstand) locomotion on diverse terrains with all gait couplings.

\subsection{Supplementary Video 2}

Hardware experiments with the Unitree Go1 and Go2. The Stage~1 and Stage~2 networks are those used throughout the paper, deployed without retraining and without modification between the two platforms; the height map that conditions both is estimated onboard by the encoder of Fig.~2 (main text) from a single forward-facing depth camera, so that no external motion capture or pre-built terrain map is used.

\textbf{Stage~1, commanded gaits.} The policy is commanded into individual gaits and executes them on demand, including a bound and a gallop across a 0.7\,m gap and over a 0.21\,m obstacle at 1\,m/s. The gap is wider than any gap presented during training in simulation. 

\textbf{Stage~2, gait selection.} The selection policy chooses and transitions between gaits online as the robot advances: walking or trotting on flat ground, and switching to a bound or gallop when approaching obstacles or gaps.

\end{document}